\documentclass[10pt]{article}
\usepackage[preprint]{tmlr}
\setcitestyle{citesep={,},yysep={,}}

\usepackage[T1]{fontenc}
\usepackage{amsmath,amssymb,amsfonts}
\usepackage{booktabs}
\usepackage{graphicx}
\usepackage{float}
\usepackage{algorithm}
\usepackage{algpseudocode}
\usepackage{needspace}
\usepackage[section]{placeins}
\usepackage{xcolor}
\usepackage{multirow}
\usepackage{array}
\usepackage[no-math]{fontspec}
\newfontfamily\figuretablefont{texgyretermes-regular.otf}[
  BoldFont=texgyretermes-bold.otf,
  ItalicFont=texgyretermes-italic.otf,
  BoldItalicFont=texgyretermes-bolditalic.otf
]
\AddToHook{env/tabular/begin}{\figuretablefont}
\usepackage{microtype}
\usepackage[hidelinks]{hyperref}
\usepackage{xurl}
\hypersetup{
  pdftitle={FinRegimeBench: Evaluation Protocols Change Conclusions in Self-Supervised Market Regime Identification},
  pdfauthor={Weijia Han and Lisha Qu},
  pdfsubject={Cross-asset market regime identification},
  pdfkeywords={self-supervised learning, financial time series, cross-asset transfer, evaluation}
}

\newcommand{\doi}[1]{doi: \begingroup\urlstyle{rm}\href{https://doi.org/#1}{\nolinkurl{#1}}\endgroup}
\newcommand{\sci}[2]{#1{\times}10^{#2}}
\title{FinRegimeBench: Evaluation Protocols Change Conclusions in
Self-Supervised Market Regime Identification\thanks{Generative AI tools
assisted with software implementation, running experiments, statistical checks,
consistency audits and manuscript editing, including prose drafts for
author-formulated interpretations. The authors verified the methods,
results and claims and take full responsibility. A fuller statement
appears before the references.}}

\author{\name Weijia Han \qquad Lisha Qu \\
      \addr University of Washington}

\def\month{November}
\def\year{2026}
\def\openreview{}

\begin{document}

\maketitle

\begin{abstract}
Identifying market regimes helps characterise changing market conditions
and informs allocation and risk decisions. Self-supervised learning offers
a way to learn market structure without predefined regime labels, but
whether its representations transfer across asset classes remains unclear.
Existing studies use different data and evaluation procedures, making it
difficult to distinguish representation quality from evaluation choices.
We introduce \textbf{\mbox{FinRegimeBench}}, a benchmark for testing whether learned
representations improve cross-asset regime identification beyond simple
correlation and volatility features. It compares methods under
shared evaluation rules and examines their sensitivity to the protocol and
market period. In the primary comparison, some methods show modest average
gains, but their lack of persistence across periods limits the evidence for
a general advantage over simple features. These findings highlight
evaluation design and temporal robustness as central to assessing progress
in financial representation learning. The benchmark provides a
reproducible basis for comparing methods and assessing which apparent
improvements are robust.
\end{abstract}

\section{Introduction}
\label{sec:intro}

Financial decisions depend on a market environment that changes over time.
A portfolio's risk reflects both the variability of individual assets and
their joint behaviour, so a change in dependence can alter the protection
provided by diversification even when the holdings remain fixed. Market
regimes describe persistent differences in this environment, including
changes in return distributions, volatility and cross-asset relationships.
The regime-switching literature gives these changes an explicit statistical
role, from latent-state models of economic fluctuations to financial models
with state-dependent risk and return~\citep{hamilton1989hmm,ang2012regime}.

These distributional changes matter to allocation and risk management.
Allocation models use estimated risk and dependence to compare portfolios, hedging decisions depend on
relationships between exposures, and risk monitoring must distinguish
persistent stress from short-lived variation. \citet{longin2001extreme}
show that international equity correlations behave differently in bear and
bull markets, with the asymmetry tied to market direction rather than
volatility alone. \citet{ang2002allocation} study portfolio choice and
currency hedging when investment opportunities change across regimes.
These findings make the structure of the market state relevant to a
decision, while leaving its economic value dependent on the investment
setting. For representation learning, the immediate requirement is to
establish that an embedding preserves useful state information before
attributing downstream value to it.

\begin{figure}[!t]
\centering
\includegraphics[width=\textwidth]{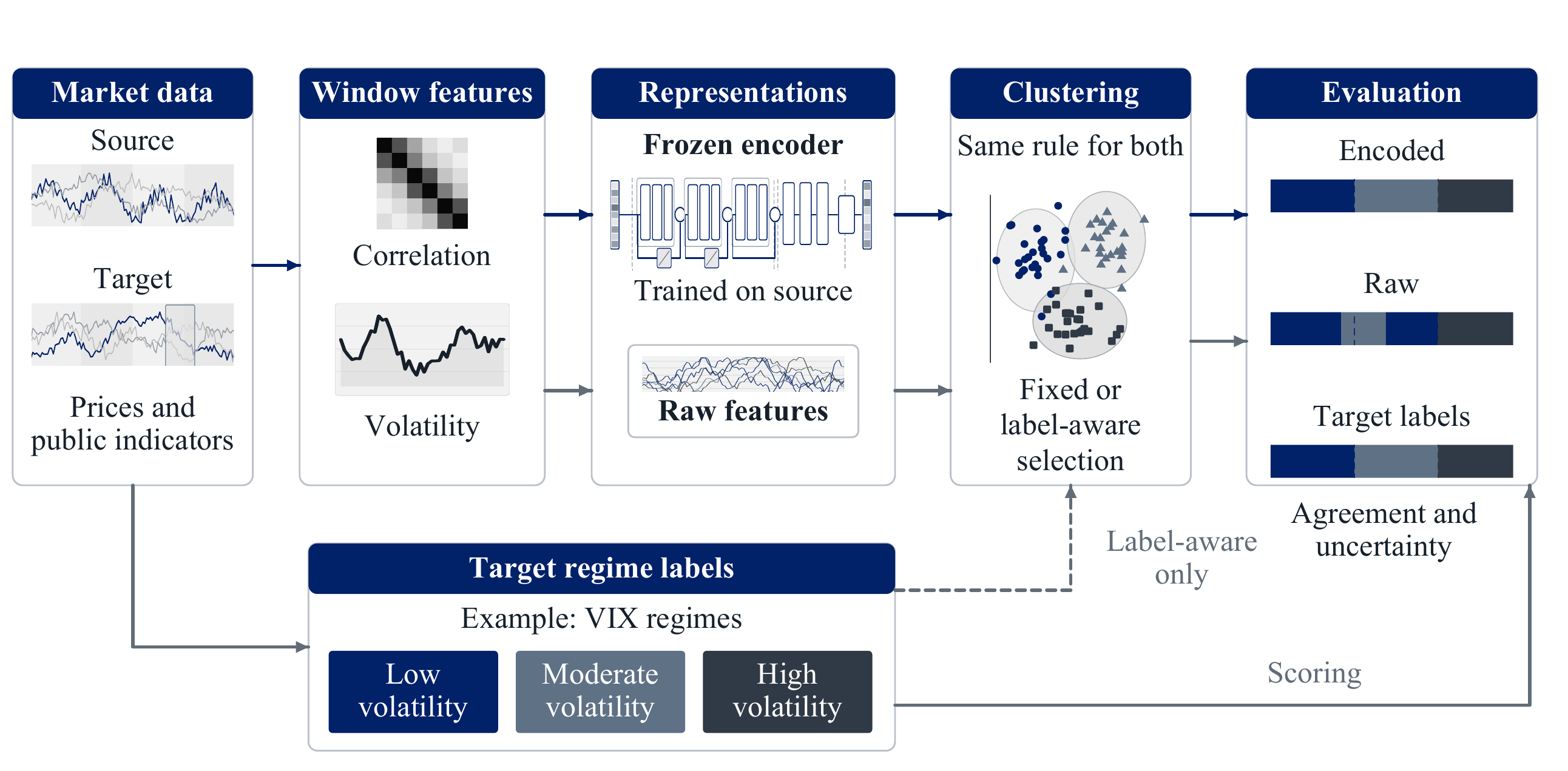}
\caption{Cross-asset regime evaluation. An encoder trained on one asset
class is frozen and applied to another. Encoded and raw features undergo
the same clustering and scoring. The dashed arrow denotes clustering
selection using target regime labels.}
\label{fig:pipeline}
\end{figure}

That requirement is difficult because regimes are latent. Markov-switching
models infer states through assumptions about conditional distributions
and transitions~\citep{hamilton1989hmm}. Persistence-penalised models place
more direct constraints on changes in state
assignments~\citep{nystrup2020learning,nystrup2021feature}. Distributional
clustering instead groups periods by the shape of their return
behaviour~\citep{horvath2024clustering}.
These approaches express different notions of a market state. Evaluation
labels introduce a related choice: a volatility threshold, a dated market
phase and a macroeconomic indicator need not identify the same periods.
Agreement with one such label rule is evidence about a particular aspect
of market structure, and the choice of rule influences what information
a learned representation is rewarded for retaining.

Self-supervised learning, or SSL, offers a way to learn structure without
using those regime assignments as training targets. Temporal objectives can
encourage nearby observations to share information, while contrastive and
regularised objectives can learn relationships that are not specified by
a label threshold~\citep{yue2022ts2vec,tonekaboni2021tnc,bardes2022vicreg}.
Financial applications provide encouraging examples with distinct goals.
\citet{ma2022denoised} use a denoising pretext task to improve labels for
price-movement classification. \citet{dolphin2024contrastive} learn asset
embeddings and evaluate them through industry classification and portfolio
optimisation. More recent work studies equity representations, downstream
prediction and their variation across market
months~\citep{merchant2026market1t}. These studies motivate learning from
financial observations, while leaving open how much transferable regime
information is captured by representations of market windows.

Cross-asset transfer makes that question especially informative. Equity,
foreign exchange and crypto differ in trading calendars, composition and
market-specific variation, yet each can exhibit persistent changes in risk
and dependence. Applying a frozen source-trained encoder to a different
asset class asks whether its representation captures structure useful
outside the market on which it was learned. To establish what learning
contributes to this transfer, the comparison needs a reference that already
describes market conditions. Simple volatility and correlation summaries
capture aspects of a regime. An encoder's value for this task
must therefore be assessed relative to those summaries, under equivalent
clustering and scoring choices. This focuses attention on the incremental
benefit of learning a representation for regime identification.
Figure~\ref{fig:pipeline} summarises this comparison and the evaluation
choices that connect each representation to a regime score.

The main obstacle is attributing a measured improvement to the representation.
A price-derived label can align closely with the volatility features used
by a baseline. A different clustering rule can change the partition
reported for an unchanged embedding. A favourable transfer direction can
also dominate an aggregate test while other directions deteriorate.
Consequently, a comparison can change because the evaluation emphasises a
different property of the data, even when the representation itself is
unchanged. This is consequential for method development: an improvement
under one arrangement may not identify which representation is more useful
under another. Benchmarking studies in other domains have similarly shown
that experimental choices influence comparative
conclusions~\citep{musgrave2020reality,bouthillier2021accounting,dehghani2021lottery}.

Financial dependence compounds this attribution problem. Overlapping
windows reuse observations, and persistent regimes can produce many labelled
windows from a small set of market episodes. Repeated training runs measure
optimisation variability on that history. Assessing another period asks
whether the learned structure remains informative under different market
conditions. Reliable evidence for transfer therefore needs both a fair
within-period comparison and an account of how its conclusion varies over
time. This motivates treating evaluation design and temporal uncertainty
as part of the research question.

We introduce FinRegimeBench to study whether learned representations improve
cross-asset regime identification over directly clustering market summaries,
and under which evaluation conditions that conclusion remains supported. The framework
compares frozen encoders and raw features on common target windows, makes
the regime-label construction explicit, and examines the influence of
selection and temporal rules. It reports average effects together with
direction-specific comparisons and variation across market periods.

\needspace{9\baselineskip}
\paragraph{Contributions.}
\begin{itemize}\setlength{\itemsep}{1pt}
\item A shared cross-asset regime-identification framework that compares
learned representations with simple market summaries under explicit data,
label and evaluation choices.
\item An empirical analysis of how label construction, clustering,
selection rules and asset universes affect the attribution of gains to
representation learning.
\item A distinction between average transfer effects, seed-level evidence
and temporal uncertainty, used to assess whether observed advantages
persist across market periods.
\end{itemize}

\section{Related Work}
\label{sec:related}

The study connects self-supervised representation learning with financial
regime identification and the evaluation practices used to compare methods.
These literatures motivate both the representations included in the
benchmark and the choices examined in its analysis.

\paragraph{Self-supervised representation learning.}
Contrastive and clustering-based objectives learn image representations
through similarity and assignment
tasks~\citep{chen2020simclr,he2020moco,caron2020swav}.
Teacher--student prediction and redundancy-reduction objectives provide
complementary routes to transferable
representations~\citep{grill2020byol,caron2021dino,zbontar2021barlow,bardes2022vicreg}.
Time-series methods adapt representation learning to temporal structure
through local neighbourhoods, multi-scale views and sequence
modelling~\citep{yue2022ts2vec,tonekaboni2021tnc,eldele2021tstcc,woo2022cost,zerveas2021transformer}.
FinRegimeBench compares representatives of these objective families on a
shared cross-asset task.

\paragraph{Foundation models for time series.}
Pretrained time-series forecasters provide another source of
representations~\citep{das2024timesfm,ansari2024chronos,woo2024moirai}.
A frozen forecaster supplies an embedding reference in our comparison,
alongside encoders trained on the benchmark's source markets.

\paragraph{Representation-learning benchmarks.}
Market-1T compares encoder-training strategies on high-frequency
US equities across multiple market conditions, with return, volatility and
spread targets~\citep{merchant2026market1t}. It examines predictive targets, embedding organisation and variation
across market months, complementing our focus on cross-asset agreement
with regime-label constructions. More
broadly, time-series representation
benchmarks vary between frozen, fine-tuned and end-to-end evaluation and
use heterogeneous classification, clustering and forecasting tasks
~\citep{trirat2026universal}. FinRegimeBench focuses on how temporal and
evaluation rules affect conclusions within a common regime-identification
task.

\paragraph{Regime identification and public labels.}
Classical regime models describe latent states and their transitions,
including Markov-switching and persistence-penalised
approaches~\citep{hamilton1989hmm,rabiner1989tutorial,nystrup2020learning,nystrup2021feature}.
\citet{ang2012regime} review their financial applications. Distributional
clustering offers a complementary perspective on recurring return
behaviour~\citep{horvath2024clustering}, while change-point and structural-break
approaches identify transitions between
periods~\citep{adams2007bocpd,killick2012pelt,bai2003multiple,vostrikova1981binary}.
Financial SSL studies instead learn representations for label construction
and asset relationships~\citep{ma2022denoised,dolphin2024contrastive}.
These approaches motivate the classical reference methods and the range of
label constructions considered here. Public macro labels include the NBER
recession indicator and the Sahm rule~\citep{sahm2019rule}. The VIX provides
an option-implied measure of market
volatility~\citep{whaley1993vix,cboe2026vix}. Comparing such constructions
helps clarify which market properties a representation's score reflects.

\paragraph{Evaluation practice.}
Evaluation choices can dominate method
differences~\citep{musgrave2020reality,bouthillier2021accounting,dehghani2021lottery}.
Our analysis studies this issue in regime identification, where temporal
dependence also affects the interpretation of repeated experiments.
Brown's method and the Kost-McDermott approximation account for dependence
when combining seed-level tests~\citep{brown1975method,kost2002combined},
with Holm correction for multiple comparisons~\citep{holm1979simple}.
We examine the question answered by these tests separately from uncertainty
across market periods.

\section{Evaluation Framework}
\label{sec:prelim}

The framework asks whether source-trained representations improve regime
identification on a different asset class. We measure their advantage over
raw features on common target windows, then examine how the comparison
changes with evaluation choices and time. Figure~\ref{fig:pipeline}
summarises the task. We first define the windows, representations and
evaluation rules, using the notation summarised in Table~\ref{tab:notation}
of Appendix~\ref{app:extra_tables}.

\paragraph{Window and transfer direction.}
A window summarises the recent state of one asset universe. The window
ending on day $u$ covers $u{-}29,\dots,u$ and produces
$x_u\in\mathbb{R}^{30}$: a volatility block contains cross-asset means of
five-day realised volatility at successive offsets, and a correlation
block contains the mean, standard deviation, 10th and 90th percentiles of
pairwise return correlations. An encoder $f_\theta$ is trained on source
windows and applied frozen to target windows. Crypto, equity and FX define
six ordered cross-asset transfer directions.

\paragraph{Evaluation block and label family.}
An evaluation block uses a common target period, with every source trained
on earlier observations. A label family assigns regime labels to one target
asset class using a fixed construction rule. It is exogenous when neither
the rule nor its data source uses target prices, and price-derived
otherwise. Section~\ref{sec:design} specifies the periods, boundary rules
and label construction.

\paragraph{Selector.}
A selector determines which clustering configuration is scored. The
candidate grid combines $k$-means and a Gaussian mixture model with
$k\in\{3,4,5\}$. The fixed selector always uses $k$-means with $k{=}3$.
The label-aware selector chooses the highest ARI against target labels
within this grid, providing a comparison of selection procedures.

\paragraph{Metrics.}
The main score is the adjusted Rand index, denoted ARI~\citep{hubert1985ari},
which compares the selected partition with the target labels over labelled
windows. The \emph{raw-feature baseline} clusters the target's own
$x_u$ under the same configurations and selector. It has no trained
parameters, so its only randomness is the clustering initialisation, and it
is scored at twenty seeds. A \emph{direction win} is a
direction in which the one-sided Welch test of the encoder's twenty
per-seed ARIs against the baseline's twenty values rejects at
$\alpha{=}0.05$. Linear-probe accuracy is reported only as a secondary
analysis in Appendix~\ref{app:probe}.

\subsection{Data and Evaluation Design}
\label{sec:design}

We first describe the market data and regime labels, then the methods and
clustering choices. The final part defines the effect sizes and separates
uncertainty across training seeds from variation across market observations.

\subsubsection{Assets, Splits and Blocks}
\label{sec:splits}
The feature universes contain twenty crypto pairs, fifty equities and
thirty FX series, selected for historical coverage. Crypto prices come from
Binance and the remaining prices from Yahoo Finance. Commodities, REITs
and bonds provide additional ETF-based families. Coverage-based tie-breaking
produces a convenience equity panel, so we examine alternative panels in
Appendix~\ref{app:universe}. Full selection rules are recorded in
Appendix~\ref{app:data_selection}.

The primary evaluation uses 2023--24 for every target, with crypto trained
on 2019--22 and equity and FX on 2015--22. We remove the first 29 target
windows to avoid returns shared with the training range and fit
price-dependent label thresholds on training data. Fixed label cutoffs
remain unchanged. Features for both learned representations and the raw
baseline are standardised using each asset's training statistics.

To assess temporal stability, we add a 2021--22 block for all three assets
and 2017--18 and 2019--20 blocks for equity and FX, with training expanded
to each block's start. The same boundary and threshold rules apply.
Table~\ref{tab:ess} provides a complementary dependence diagnostic using
the window counts from the earlier mixed-horizon setting described below.

Some secondary experiments use an earlier mixed-horizon setting: crypto is
evaluated on 2023--24 and equity and FX on 2021--24. This allows calendar
overlap between crypto-source training and target evaluation. We identify
these analyses separately and connect them to the primary setting through
the protocol ablation in Section~\ref{sec:protocol}. Held-out commodities
use separate validation and test periods, as described in Appendix~\ref{app:heldout}.

\subsubsection{Label Families}
\label{sec:labels}

\begin{table}[t]
\caption{Regime-label families. E and P denote exogenous and price-derived
labels. Align ARI compares the labels with $k{=}3$ $k$-means on volatility
features alone. Paired FX entries use thresholds fitted through 2020 and
2022, respectively, and paired commodity entries denote validation and
test. Manual-crypto Classes gives overall and evaluation-period counts.
The first three families form the primary aggregate.}
\label{tab:families}
\centering
\footnotesize
\setlength{\tabcolsep}{4pt}
\begin{tabular}{@{}llp{7.4cm}ccc@{}}
\toprule
Family & Target & States / short rule & Classes & Flag & Align ARI \\
\midrule
Manual crypto & crypto & Author-dated market phases. The evaluation range contains macro-stress and recovery periods. & 4 / 2 & P & $-0.019$ \\
VIX-level & equity & VIX close below 15, from 15 to 25, or at least 25 & 3 & P & $0.102$ \\
FX volatility & FX & Training-period quartiles of 30-day realised FX volatility & 3 & P & $0.325$ / $0.186$ \\
\midrule
VIX-level, shifted & equity & VIX cutoffs 13/23 or 17/27 are defined but not scored. & 3 & P & n/a \\
Sahm rule & equity & Published Sahm value below 0.5, from 0.5 to 1.0, or at least 1.0 & 3 & E & $-0.022$ \\
NBER recession & equity & The NBER recession indicator is single-class in the evaluation range. & 2 & E & n/a \\
Commodities vol. & commodities & Training-period quartiles of ETF realised volatility & 3 & P & $0.078$ / $0.255$ \\
REIT vol. & REITs & Training-period quartiles of ETF realised volatility & 3 & P & $0.278$ \\
Bond vol. & bonds & Training-period quartiles of ETF realised volatility & 3 & P & $0.370$ \\
\bottomrule
\end{tabular}
\end{table}

The first three families in Table~\ref{tab:families} form the primary
aggregate, matching each target to its label source. Manual crypto and
VIX-level labels use fixed rules. Realised-volatility labels use
training-period percentile cutoffs. The alignment column measures how
closely the labels agree with a clustering of volatility features alone.
This helps interpret performance when representations and label rules
capture related aspects of market variation.

\subsubsection{Methods}
\label{sec:methods}
The primary comparison keeps the market summaries fixed and varies how target
representations or state assignments are constructed.
Algorithm~\ref{alg:method_groups} separates the primary encoders,
sensitivity analyses and reference constructions. The six joint-embedding
objectives form $\mathcal J$ and share a residual MLP and optimisation
schedule, pairing anchors with augmented temporal neighbours. The three
convolutional sequence methods form $\mathcal Q$.

Write $X_s^{\mathrm{tr}}$, $X_s^{\mathrm{ev}}$ and
$X_t^{\mathrm{ev}}=\{x_{t,u}\}_u$ for source training, source evaluation
and scored target windows. Each $\operatorname{Fit}_m$ uses the specified
method's objective and training procedure, with separate parameters.
Each $f_{\theta_m}$ includes the method's embedding extraction rule.
All trained parameters are frozen for target embedding.

\begin{algorithm}[H]
\caption{Representation and state construction by method group}
\label{alg:method_groups}
\small
\figuretablefont
\algrenewcommand{\algorithmicrequire}{\textbf{Input:}}
\algrenewcommand{\algorithmicensure}{\textbf{Output:}}
\begin{algorithmic}[1]
\Require $X_s^{\mathrm{tr}},X_t^{\mathrm{ev}}$, method specifications and row-specific $X_s^{\mathrm{ev}},C_t^{(m)},y_s^{\mathrm{tr}},R_{t,u}$.
\Ensure Embeddings $Z_t^{(m)}$ or state assignments $\hat c_t^{(m)}$, grouped as A, B or C.
\Statex \textbf{A. Primary window-wise encoders}
\ForAll{$m\in\mathcal J$}
  \State $\theta_m\gets\operatorname{Fit}_m(X_s^{\mathrm{tr}})$
    \Comment{shared residual MLP}
\EndFor
\State $(\theta_{\mathrm{TNC\text{-}style}},\phi_{\mathrm{TNC\text{-}style}})\gets\operatorname{Fit}_{\mathrm{TNC\text{-}style}}(X_s^{\mathrm{tr}})$
  \Comment{plain MLP, neighbour discriminator}
\State $Z_t^{(m)}[u]\gets f_{\theta_m}(x_{t,u})$
  \Comment{$m\in\mathcal J\cup\{\mathrm{TNC\text{-}style}\}$, window-wise}
\Statex \textbf{B. Sensitivity constructions}
\ForAll{$m\in\mathcal Q$}
  \State $\theta_m\gets\operatorname{Fit}_m(X_s^{\mathrm{tr}})$
    \Comment{source sequence}
  \State $Z_t^{(m)}[u]\gets f_{\theta_m}(x_{t,u}\mid C_t^{(m)})$
    \Comment{target-sequence context}
\EndFor
\State $\{\theta_e\}_e\gets\operatorname{Fit}_{\mathrm{VAE}}(X_s^{\mathrm{tr}})$
  \Comment{epoch checkpoints}
\State $e^\star\gets\arg\min_e\mathcal L_{\mathrm{VAE}}(\theta_e,X_s^{\mathrm{ev}})$
  \Comment{unlabelled source evaluation}
\State $Z_t^{\mathrm{VAE}}[u]\gets\mu_{\theta_{e^\star}}(x_{t,u})$
  \Comment{posterior mean}
\Statex \textbf{C. Reference constructions}
\State $P_s\gets\operatorname{PCAFit}(X_s^{\mathrm{tr}}),\quad Z_t^{\mathrm{PCA}}\gets P_s(X_t^{\mathrm{ev}})$
  \Comment{source-fitted projection}
\State $Z_t^{(b)}[u]\gets f_{\theta_b^0}(x_{t,u})$
  \Comment{untrained backbone $b$}
\State $\hat c_t^{(m)}\gets\operatorname{FitStates}_m(X_t^{\mathrm{ev}})$
  \Comment{HMM or jump model}
\State $\mathcal S_t^{(m)}\gets\operatorname{Segment}_m(X_t^{\mathrm{ev}})$
  \Comment{PELT or binary segmentation}
\State $\hat c_t^{(m)}\gets\operatorname{StateMap}(\mathcal S_t^{(m)},X_t^{\mathrm{ev}})$
  \Comment{cluster segment means}
\Statex \hspace{\algorithmicindent}\textit{Earlier mixed-horizon setting only}
\State $\theta_{\mathrm{sup}}\gets\operatorname{Fit}_{\mathrm{CE}}(X_s^{\mathrm{tr}},y_s^{\mathrm{tr}})$
  \Comment{source proxy labels}
\State $Z_t^{\mathrm{sup}}[u]\gets f_{\theta_{\mathrm{sup}}}(x_{t,u})$
  \Comment{supervised reference}
\State $Z_t^{\mathrm{pre}}[u]\gets\operatorname{Pool}(F_{\mathrm{pre}}(R_{t,u}))$
  \Comment{pretrained returns model}
\end{algorithmic}
\end{algorithm}

The context $C_t^{(m)}$ contains other target windows, including excluded
boundary windows and complete-block information where applicable.
$\mathcal L_{\mathrm{VAE}}$ includes reconstruction and KL terms.
For the earlier references, $y_s^{\mathrm{tr}}$ denotes source proxy labels
and $R_{t,u}$ a target return window for the frozen forecaster
$F_{\mathrm{pre}}$~\citep{ansari2024chronos}.
Appendix~\ref{app:methods} details the objectives, architectures and
extraction rules, and Appendix~\ref{app:faithful} examines
published-objective variants. Embeddings enter the selectors below.
Direct state assignments enter scoring.

\subsubsection{Selectors}
A learned embedding defines a geometry, but the reported regime partition
also depends on how that geometry is clustered and how a configuration is
selected. The primary selector is fixed $k{=}3$ $k$-means,
motivated by the three-state equity and FX labels. Other label-free
selectors use fixed GMM clustering or choose from the grid using
silhouette~\citep{rousseeuw1987silhouette},
Calinski--Harabasz~\citep{calinski1974dendrite} or
Davies--Bouldin~\citep{davies1979cluster} scores. Label-aware selection
instead maximises agreement with evaluation labels. Applying each rule to
the same stored grids for encoders and raw features isolates selection
sensitivity without retraining.

\subsubsection{Inference}
\label{sec:inference}
The main effect is the difference between an encoder's seed-mean ARI and
the raw-feature baseline's seed-mean ARI on the same target windows, averaged across the available
transfer directions. Direction-level effects accompany this average to
show where any gain arises.

\paragraph{Temporal uncertainty.}
We use two complementary summaries. Comparing effects across evaluation
blocks describes how the result changes over market periods.
A circular moving-block bootstrap~\citep{kunsch1989jackknife,politis1992circular}
resamples dependent windows within a period and refits clustering, giving
intervals for individual direction-level effects. Neither procedure supplies
new regime histories. Episode counts and first-order autoregressive effective sample sizes
characterise dependence within each period. The full calculations and
bootstrap settings are in Appendix~\ref{app:inference}.

\paragraph{Seed-level tests.}
A one-sided Welch test~\citep{welch1947generalization} compares the twenty
per-seed ARIs for each encoder and direction. A direction win is a rejection
at $\alpha{=}0.05$. Encoder runs and raw clustering runs use independent random draws. Their
seeds vary training and clustering initialisation, respectively. We combine direction-level
tests using Fisher's statistic~\citep{fisher1925statistical} with the
Brown/Kost--McDermott dependence correction~\citep{brown1975method,kost2002combined}.
The resulting $p_a$ can be small because of a favourable direction even
when the average effect is negative. We therefore report it alongside
effect sizes and direction-win counts. Holm correction~\citep{holm1979simple}
applies separately to each selector over the seven primary encoders.
Source-independent baselines and sensitivity rows are reported separately.
Appendix~\ref{app:calibration} examines calibration and
Appendix~\ref{app:pairing} examines row pairing and covariance sensitivity.

\section{Results}
\label{sec:findings}

We first ask whether learned representations improve on raw features in
the primary evaluation. We then examine what the statistical tests detect,
which evaluation choices change the comparison, and whether the effects
persist across market periods. Additional methods and tasks extend the
analysis beyond this central comparison. Unless stated otherwise, results
refer to the primary evaluation. Significance means that a seed-level
result survives Holm correction within its stated comparison family.

\Needspace{23\baselineskip}
\subsection{Primary Effects}
\label{sec:raw}

\begin{table}[H]
\caption{Primary comparison with raw features. Mean ARI averages seeds and
six transfer directions under fixed $k{=}3$ $k$-means. $\Delta$ is the
corresponding encoder-minus-raw difference. Wins counts one-sided seed-level
Welch rejections at $\alpha{=}0.05$. $p_a$ is the Brown-adjusted Fisher $p$
across directions. $^{\dagger}$ marks Holm significance among the seven
primary encoders, separately for each selector. Sensitivity rows are
outside this family.}
\label{tab:clean_main}
\centering
\small
\setlength{\tabcolsep}{5.5pt}
\begin{tabular}{@{}lccccc@{}}
\toprule
Method & Mean ARI & $\Delta$ & Wins / 6 & Fixed $p_a$ & Label-aware $p_a$ \\
\midrule
VICReg-TN & 0.092 & $+0.022$ & 4 & $<0.001$\,$^{\dagger}$ & $<0.001$\,$^{\dagger}$ \\
SimCLR+VC & 0.088 & $+0.018$ & 4 & $<0.001$\,$^{\dagger}$ & $<0.001$\,$^{\dagger}$ \\
Barlow Twins & 0.046 & $-0.024$ & 2 & $<0.001$\,$^{\dagger}$ & $<0.001$\,$^{\dagger}$ \\
BYOL & 0.003 & $-0.067$ & 1 & $0.577$ & $0.001$\,$^{\dagger}$ \\
BT+VH & 0.045 & $-0.025$ & 2 & $<0.001$\,$^{\dagger}$ & $<0.001$\,$^{\dagger}$ \\
VICReg-Hard & 0.004 & $-0.066$ & 0 & $1.000$ & $0.065$ \\
TNC-style & 0.063 & $-0.007$ & 1 & $<0.001$\,$^{\dagger}$ & $<0.001$\,$^{\dagger}$ \\
\midrule
\multicolumn{6}{@{}l}{\emph{Sensitivity rows with evaluation-time exceptions}} \\
TS2Vec-style & 0.082 & $+0.012$ & 4 & $<0.001$ & $<0.001$ \\
TS-TCC-style & 0.023 & $-0.047$ & 0 & $0.998$ & $0.944$ \\
CoST-style & 0.033 & $-0.037$ & 1 & $0.188$ & $0.021$ \\
VAE & 0.051 & $-0.019$ & 2 & $0.066$ & $<0.001$ \\
\midrule
Baseline, $k{=}3$ $k$-means & 0.070 &  &  &  &
 \\
\bottomrule
\end{tabular}
\end{table}

VICReg-TN and SimCLR+VC show modest positive average effects, while the
remaining primary encoders fall below raw on average, as shown in
Table~\ref{tab:clean_main}. Several negative-average rows nevertheless
have significant combined tests. Interpreting this pattern requires a
closer look at how an average effect differs from the evidence aggregated
by the seed-level statistic.

\FloatBarrier
\subsection{What the Seed-Level Test Detects}
\label{sec:estimand}

The average effect gives equal weight to each direction's ARI difference.
The combined test instead aggregates one-sided evidence for improvement
within individual directions. Strong seed-level evidence in a favourable
direction can therefore produce a rejection even when losses elsewhere make
the average negative. The distribution of the direction-level effects
explains how the two summaries can lead to different interpretations of
the same comparison.

The calibration study in Appendix~\ref{app:calibration} examines the
statistic's size and power under the fitted seed-dependence model.
Calibration addresses its behaviour for that question, while the average
effect remains the measure of overall improvement used here. Both summaries
also depend on the evaluation procedure that produces their scores. This
motivates examining the boundary and label rules before asking whether an
advantage persists across market periods.

\subsection{Sensitivity to the Evaluation Procedure}
\label{sec:protocol}

We vary boundary exclusion and label-threshold fitting separately to
examine how evaluation rules change the apparent advantage of learning.

\begin{figure}[H]
\centering
\includegraphics[width=\textwidth]{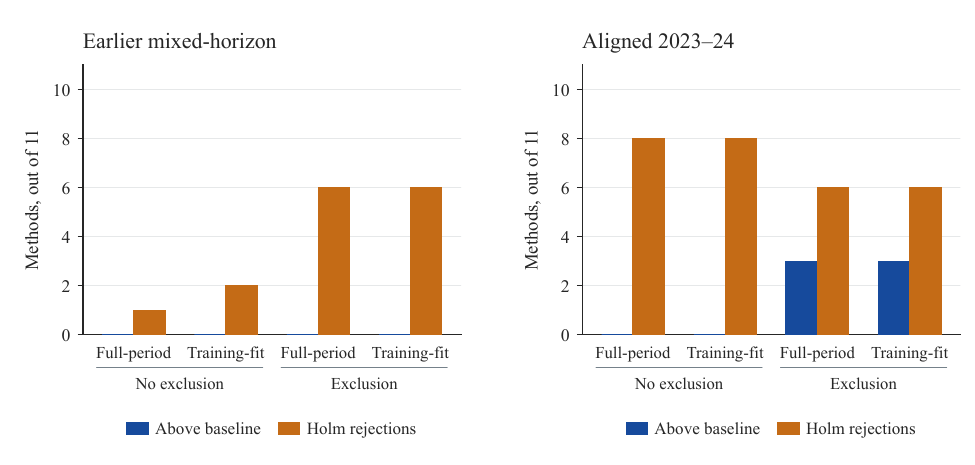}
\caption{Protocol sensitivity across eleven methods under fixed $k{=}3$
$k$-means. Bars count methods with positive six-direction mean ARI
differences from raw and Holm-corrected combined seed-test rejections.
The panels vary boundary exclusion and data-dependent label-threshold
fitting within each evaluation setting. Full results are in
Table~\ref{tab:protocol_ablation}.}
\label{fig:protocol_effect_counts}
\end{figure}

Seed-level rejections can occur even when no method has a positive
average effect. The full ablation in Table~\ref{tab:protocol_ablation} shows that
removing overlapping boundary windows changes the raw-feature baseline
more than the learned representations, especially on crypto. Positive average effects in the primary period appear only after
this exclusion. Training-only threshold fitting also changes absolute
scores, but has less influence on the direction of the average comparison.

The same sensitivity appears in the earlier mixed-horizon setting.
Together, the ablations show how temporal separation can change the
reference baseline and hence the apparent benefit of learning.

\FloatBarrier
\subsection{Stability Across Evaluation Blocks}
\label{sec:ess}
\label{sec:folds}

We now retain the same boundary and threshold rules while changing the
market block and preceding training history. This comparison asks whether
the observed advantages persist across market periods.

\begin{figure}[H]
\centering
\includegraphics[width=\textwidth]{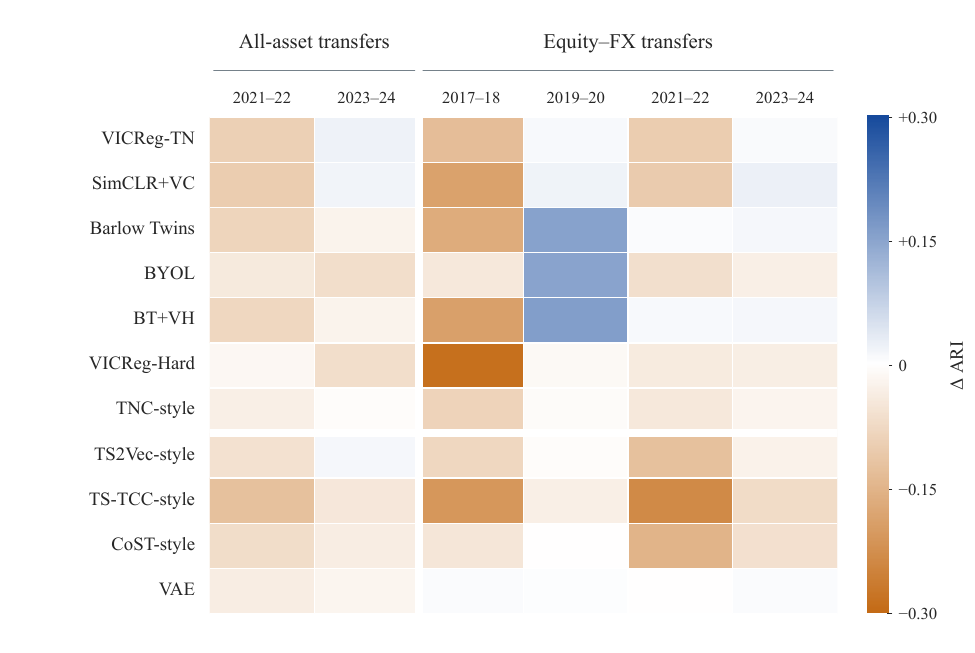}
\caption{Transfer effects across market periods under fixed $k{=}3$
$k$-means. Colour shows seed-mean ARI differences from raw, averaged over
six directions in the all-asset columns and two directions in the equity--FX
columns. The horizontal divider
separates primary encoders from sensitivity rows. Full values and
across-period means are in Tables~\ref{tab:period} and~\ref{tab:period_equityfx}.}
\label{fig:temporal_effects}
\end{figure}

The positive primary-period effects do not persist across the available
blocks, as shown in Figure~\ref{fig:temporal_effects}.
Several methods change sign across the equity--FX periods. Differences
in market conditions, label episodes and baseline performance accompany
these changes. The observed advantage is consequently specific to the
period as well as the representation. This variation across periods leaves
open how much uncertainty is already present within an individual block.
Resampling contiguous windows examines that question while retaining
local temporal dependence.

\FloatBarrier
\subsection{Uncertainty Within an Evaluation Block}
\label{sec:bootstrap}

\begin{figure}[H]
\centering
\includegraphics[width=0.80\textwidth]{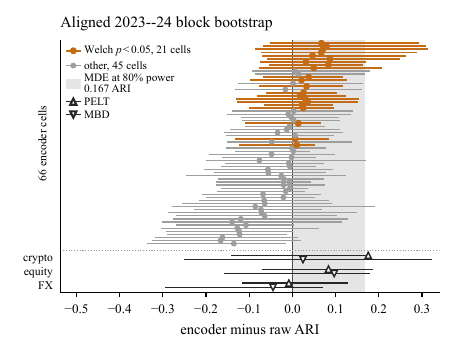}
\caption{ARI differences from raw features in the aligned 2023--24
evaluation. Points show estimated effects and lines show 95\% moving-block
bootstrap intervals. Orange marks seed-level wins. Shading indicates the
minimum detectable positive effect at 80\% power, using the median
bootstrap standard error.}
\label{fig:bootstrap}
\end{figure}

The moving-block analysis complements the period comparisons by varying
the dependent window sequence within a fixed evaluation block.
Figure~\ref{fig:bootstrap} shows the resulting direction-level comparisons,
which have substantially more uncertainty than repeated training seeds. Every primary-evaluation interval crosses zero
when a direction has a seed-level win, and none lies wholly above zero.
Thus, the precise seed-level comparisons do not establish similarly precise
gains across possible window sequences.

These direction-specific intervals complement the comparison of average
effects across periods. Together, they limit the evidence for a stable
method ranking. Appendix~\ref{app:classical} gives the corresponding
values and classical-detector results.

\FloatBarrier
\subsection{Complementary Analyses}

\paragraph{Representation and selection checks.}
\label{sec:flip}
\label{sec:kselector}
\label{sec:controls}
\label{sec:baselines}
\label{sec:tuning}
The primary analyses establish the scale and temporal uncertainty of the
observed gains. Complementary comparisons examine how that pattern relates
to the representation, its readout and the evaluation target.

Projection and untrained-network controls reach the range of several
trained encoders in the earlier setting, indicating a role for compression,
architecture and embedding geometry. Clustering choices can also change
the apparent advantage through movement in the raw baseline, and
label-aware selection generally produces more rejections. Objective
variants, classical detectors and label-free tuning alter individual
outcomes without establishing a stable advantage. One sequence-based
sensitivity row has a positive primary-period average under its different
evaluation procedure. Alternative equity panels reduce the averages of
the positive-average primary rows without reversing them. These checks
retain their distinct information-access and universe conditions, as
reported in Appendices~\ref{app:controls}, \ref{app:faithful},
\ref{app:classical}, \ref{app:universe} and~\ref{app:extra_tables}.

\paragraph{Additional label families and probes.}
\label{sec:heldout}
Changing the label family examines a different source of sensitivity:
which market property is rewarded by the score. The exogenous Sahm labels
yield near-chance partitions for learned and raw representations, despite
significant relative margins. Price-based commodity, REIT and bond labels
show stronger effects in some settings. Linear probes provide a further
check by changing the readout from clustering to supervised prediction,
with isolated gains. The periods and checkpoint procedures underlying
these comparisons are documented in Appendices~\ref{app:sahm},
\ref{app:heldout}, \ref{app:families} and~\ref{app:probe}.

\paragraph{Synthetic study.}
\label{sec:synthetic}
Known-regime simulations complement the proxy-label evidence by making
the state-generating mechanism explicit. Contrasting changes in feature
levels with changes in dynamics examines which distinctions the market
summaries expose to learned representations and raw clustering. Raw
features recover level changes more readily, while comparative performance
varies with the objective and selector. The study illustrates these
mechanisms on one generated series per condition
as described in Appendix~\ref{app:synthetic}.

\section{Discussion}
\label{sec:limits}
The results place representation gains in the context of the market
property being measured. A gain depends on the labels, the geometry used
for clustering and the market history over which it is assessed. The
raw-feature comparisons and controls therefore motivate identifying when
learning improves regime identification over direct clustering of simple
risk and dependence summaries. This connects method development to the market structure a
representation is intended to preserve.

The primary comparison examines offline clustering of fixed summaries against
proxy labels, so agreement provides evidence about the property described
by each label rule. Broader exogenous label coverage and greater episode
diversity would extend that evidence, particularly for crypto, where the
evaluation contains few regime episodes. Matched comparisons using raw return
sequences and broader pretrained models would explore the role of richer
inputs and prior learning. Online detection and downstream allocation or
risk tasks would then examine operational relevance beyond proxy-label
agreement.

Interpretation also benefits from keeping methodological choices visible.
Primary encoders use method-specific embedding outputs and extraction
rules, and dropout remains active in the teacher--student target.
Sequence context and the reconstruction model's source-evaluation
checkpoint rule motivate the separate sensitivity group. Several secondary analyses retain
mixed-horizon checkpoints with calendar overlap, and their findings apply
to that evaluation setting. Matched extraction and information-access
policies would help isolate the contributions of objectives,
architectures and evaluation rules.

Data composition offers a complementary direction for replication.
Retrospective universes inherit survivorship and coverage effects, and
the convenience equity panel affects target-side comparisons. The
current universe checks examine that target-side sensitivity.
Point-in-time universes, together with retraining across representative
source panels, would extend the analysis to variation in both training
and target populations.

Temporal replication is especially important for interpreting transfer.
The available blocks are not exchangeable market samples, so their
differences describe observed historical variation. Moving-block
intervals complement this view at the direction level, conditional on
block length and raw-clustering initialisation. Inference for the
cross-direction mean across market histories calls for further analysis
that accounts for shared sources and targets. The synthetic training
seeds also share one generated series per condition. Additional market
periods and independently generated series would expand the evidence for
assessing whether an advantage persists across regime histories.

\Needspace{14\baselineskip}
\section{Conclusion}
\label{sec:conclusion}
FinRegimeBench provides a shared framework for assessing whether learned
representations improve cross-asset market regime identification beyond
simple correlation and volatility features. It combines common evaluation
rules with diagnostic controls that relate comparative performance to
label definitions, clustering choices and market history. The analysis
distinguishes evidence across training seeds from evidence across market
periods. The modest gains in the primary evaluation do not persist across
the available periods, underscoring the importance of assessing both
average effects and temporal robustness. Together, these comparisons offer
a reproducible basis for evaluating progress in transferable financial
representations.

Future work can extend the benchmark to more diverse market histories and
exogenous regime labels, alongside representations learned from richer
return sequences. Evaluating online regime detection and downstream risk
tasks would further test the relevance of these
representations to practical decisions.

\Needspace{9\baselineskip}
\subsection*{Reproducibility statement}
The paper and appendices specify the data definitions, temporal splits,
label-fitting rules, model configurations and statistical procedures used
in the study. The descriptions distinguish analyses based on recorded
per-seed scores from those requiring embeddings or feature matrices.
Public data sources and
licence information are listed in Appendix~\ref{app:data}. Experimental settings and resource use are described in
Appendices~\ref{app:methods} and~\ref{app:compute}.

\subsection*{AI use statement}
Generative AI tools assisted with software implementation, running
experiments, statistical checks, consistency audits and prose editing.
The authors formulated the interpretations and defined the question,
targets, methods and tests. They verified the software and results,
regenerated all tables and reviewed the manuscript and figures.
Market data were not generated by AI.
The authors take full responsibility.

\bibliographystyle{tmlr}
\bibliography{references}

\begin{thebibliography}{51}
\providecommand{\natexlab}[1]{#1}
\providecommand{\url}[1]{\texttt{#1}}
\expandafter\ifx\csname urlstyle\endcsname\relax
  \providecommand{\doi}[1]{doi: #1}\else
  \providecommand{\doi}{doi: \begingroup \urlstyle{rm}\Url}\fi

\bibitem[Adams \& MacKay(2007)Adams and MacKay]{adams2007bocpd}
Ryan~Prescott Adams and David J.~C. MacKay.
\newblock {B}ayesian online changepoint detection.
\newblock \emph{arXiv preprint arXiv:0710.3742}, 2007.
\newblock URL \url{https://arxiv.org/abs/0710.3742}.

\bibitem[Ang \& Bekaert(2002)Ang and Bekaert]{ang2002allocation}
Andrew Ang and Geert Bekaert.
\newblock International asset allocation with regime shifts.
\newblock \emph{The Review of Financial Studies}, 15\penalty0 (4):\penalty0
  1137--1187, 2002.
\newblock \doi{10.1093/rfs/15.4.1137}.
\newblock URL
  \url{https://business.columbia.edu/sites/default/files-efs/imce-uploads/gbekaert/papers/international%20assel%20allocation.pdf}.

\bibitem[Ang \& Timmermann(2012)Ang and Timmermann]{ang2012regime}
Andrew Ang and Allan Timmermann.
\newblock Regime changes and financial markets.
\newblock \emph{Annual Review of Financial Economics}, 4\penalty0 (1):\penalty0
  313--337, 2012.
\newblock \doi{10.1146/annurev-financial-110311-101808}.
\newblock URL
  \url{https://www.annualreviews.org/content/journals/10.1146/annurev-financial-110311-101808}.

\bibitem[Ansari et~al.(2024)Ansari, Stella, Turkmen, Zhang, Mercado, Shen,
  Shchur, Rangapuram, Pineda~Arango, Kapoor, et~al.]{ansari2024chronos}
Abdul~Fatir Ansari, Lorenzo Stella, Caner Turkmen, Xiyuan Zhang, Pedro Mercado,
  Huibin Shen, Oleksandr Shchur, Syama~Sundar Rangapuram, Sebastian
  Pineda~Arango, Shubham Kapoor, et~al.
\newblock Chronos: Learning the language of time series.
\newblock \emph{Transactions on Machine Learning Research (TMLR)}, 2024.
\newblock URL \url{https://arxiv.org/abs/2403.07815}.

\bibitem[Bai \& Perron(2003)Bai and Perron]{bai2003multiple}
Jushan Bai and Pierre Perron.
\newblock Computation and analysis of multiple structural change models.
\newblock \emph{Journal of Applied Econometrics}, 18\penalty0 (1):\penalty0
  1--22, 2003.
\newblock \doi{10.1002/jae.659}.
\newblock URL
  \url{https://www.columbia.edu/~jb3064/papers/2003_Computation_and_analysis_of_multiple_structural_changes.pdf}.

\bibitem[Bardes et~al.(2022)Bardes, Ponce, and LeCun]{bardes2022vicreg}
Adrien Bardes, Jean Ponce, and Yann LeCun.
\newblock {VICReg}: Variance-invariance-covariance regularization for
  self-supervised learning.
\newblock In \emph{International Conference on Learning Representations
  (ICLR)}, 2022.
\newblock URL \url{https://arxiv.org/abs/2105.04906}.

\bibitem[Bouthillier et~al.(2021)Bouthillier, Delaunay, Bronzi, Trofimov,
  Nichyporuk, Szeto, Mohammadi~Sepahvand, Raff, Madan, Voleti, Ebrahimi~Kahou,
  Michalski, Serdyuk, Arbel, Pal, Varoquaux, and
  Vincent]{bouthillier2021accounting}
Xavier Bouthillier, Pierre Delaunay, Mirko Bronzi, Assya Trofimov, Brennan
  Nichyporuk, Justin Szeto, Nazanin Mohammadi~Sepahvand, Edward Raff, Kanika
  Madan, Vikram Voleti, Samira Ebrahimi~Kahou, Vincent Michalski, Dmitriy
  Serdyuk, Tal Arbel, Chris Pal, Ga{\"e}l Varoquaux, and Pascal Vincent.
\newblock Accounting for variance in machine learning benchmarks.
\newblock In \emph{Proceedings of Machine Learning and Systems (MLSys)},
  volume~3, pp.\  747--769, 2021.
\newblock URL
  \url{https://proceedings.mlsys.org/paper_files/paper/2021/hash/0184b0cd3cfb185989f858a1d9f5c1eb-Abstract.html}.

\bibitem[Brown(1975)]{brown1975method}
Morton~B. Brown.
\newblock A method for combining non-independent, one-sided tests of
  significance.
\newblock \emph{Biometrics}, 31\penalty0 (4):\penalty0 987--992, 1975.
\newblock \doi{10.2307/2529826}.
\newblock URL \url{https://www.jstor.org/stable/2529826}.

\bibitem[Cali{\'n}ski \& Harabasz(1974)Cali{\'n}ski and
  Harabasz]{calinski1974dendrite}
Tadeusz Cali{\'n}ski and J.~Harabasz.
\newblock A dendrite method for cluster analysis.
\newblock \emph{Communications in Statistics}, 3\penalty0 (1):\penalty0 1--27,
  1974.
\newblock \doi{10.1080/03610927408827101}.
\newblock URL
  \url{https://www.researchgate.net/profile/Tadeusz-Calinski/publication/233096619_A_dendrite_method_for_cluster_analysis/links/555213e108aeaaff3befe29b/A-dendrite-method-for-cluster-analysis.pdf}.

\bibitem[Caron et~al.(2020)Caron, Misra, Mairal, Goyal, Bojanowski, and
  Joulin]{caron2020swav}
Mathilde Caron, Ishan Misra, Julien Mairal, Priya Goyal, Piotr Bojanowski, and
  Armand Joulin.
\newblock Unsupervised learning of visual features by contrasting cluster
  assignments.
\newblock In \emph{Advances in Neural Information Processing Systems
  (NeurIPS)}, volume~33, pp.\  9912--9924, 2020.
\newblock URL
  \url{https://proceedings.neurips.cc/paper/2020/hash/70feb62b69f16e0238f741fab228fec2-Abstract.html}.

\bibitem[Caron et~al.(2021)Caron, Touvron, Misra, J{\'e}gou, Mairal,
  Bojanowski, and Joulin]{caron2021dino}
Mathilde Caron, Hugo Touvron, Ishan Misra, Herv{\'e} J{\'e}gou, Julien Mairal,
  Piotr Bojanowski, and Armand Joulin.
\newblock Emerging properties in self-supervised vision transformers.
\newblock In \emph{Proceedings of the IEEE/CVF International Conference on
  Computer Vision (ICCV)}, pp.\  9650--9660, 2021.
\newblock URL \url{https://arxiv.org/abs/2104.14294}.

\bibitem[{Cboe Global Indices, LLC}(2026)]{cboe2026vix}
{Cboe Global Indices, LLC}.
\newblock {Cboe} volatility index methodology.
\newblock
  \url{https://cdn.cboe.com/resources/indices/governance/Volatility_Index_Methodology_Cboe_Volatility_Index.pdf},
  2026.
\newblock Version 6.0, revised February 26, 2026.

\bibitem[Chen et~al.(2020)Chen, Kornblith, Norouzi, and Hinton]{chen2020simclr}
Ting Chen, Simon Kornblith, Mohammad Norouzi, and Geoffrey Hinton.
\newblock A simple framework for contrastive learning of visual
  representations.
\newblock In \emph{International Conference on Machine Learning (ICML)}, pp.\
  1597--1607, 2020.
\newblock URL \url{https://proceedings.mlr.press/v119/chen20j.html}.

\bibitem[Das et~al.(2024)Das, Kong, Sen, and Zhou]{das2024timesfm}
Abhimanyu Das, Weihao Kong, Rajat Sen, and Yichen Zhou.
\newblock A decoder-only foundation model for time-series forecasting.
\newblock In \emph{International Conference on Machine Learning (ICML)}, volume
  235, pp.\  10148--10167, 2024.
\newblock URL \url{https://proceedings.mlr.press/v235/das24c.html}.

\bibitem[Davies \& Bouldin(1979)Davies and Bouldin]{davies1979cluster}
David~L. Davies and Donald~W. Bouldin.
\newblock A cluster separation measure.
\newblock \emph{IEEE Transactions on Pattern Analysis and Machine
  Intelligence}, PAMI-1\penalty0 (2):\penalty0 224--227, 1979.
\newblock \doi{10.1109/TPAMI.1979.4766909}.
\newblock URL
  \url{https://www.researchgate.net/profile/Don_Bouldin/publication/224377470_A_Cluster_Separation_Measure/links/02e7e52d57df1ad121000000.pdf}.

\bibitem[Dehghani et~al.(2021)Dehghani, Tay, Gritsenko, Zhao, Houlsby, Diaz,
  Metzler, and Vinyals]{dehghani2021lottery}
Mostafa Dehghani, Yi~Tay, Alexey~A. Gritsenko, Zhe Zhao, Neil Houlsby, Fernando
  Diaz, Donald Metzler, and Oriol Vinyals.
\newblock The benchmark lottery.
\newblock \emph{arXiv preprint arXiv:2107.07002}, 2021.
\newblock URL \url{https://arxiv.org/abs/2107.07002}.

\bibitem[Dolphin et~al.(2024)Dolphin, Smyth, and Dong]{dolphin2024contrastive}
Rian Dolphin, Barry Smyth, and Ruihai Dong.
\newblock Contrastive learning of asset embeddings from financial time series.
\newblock In \emph{Proceedings of the Fifth ACM International Conference on AI
  in Finance (ICAIF)}, 2024.
\newblock \doi{10.1145/3677052.3698610}.
\newblock URL \url{https://arxiv.org/abs/2407.18645}.

\bibitem[Eldele et~al.(2021)Eldele, Ragab, Chen, Wu, Kwoh, Li, and
  Guan]{eldele2021tstcc}
Emadeldeen Eldele, Mohamed Ragab, Zhenghua Chen, Min Wu, Chee~Keong Kwoh,
  Xiaoli Li, and Cuntai Guan.
\newblock Time-series representation learning via temporal and contextual
  contrasting.
\newblock In \emph{Proceedings of the 30th International Joint Conference on
  Artificial Intelligence (IJCAI)}, pp.\  2352--2359, 2021.
\newblock \doi{10.24963/ijcai.2021/324}.
\newblock URL \url{https://ijcai.org/proceedings/2021/324}.

\bibitem[Fisher(1925)]{fisher1925statistical}
Ronald~A. Fisher.
\newblock \emph{Statistical Methods for Research Workers}.
\newblock Oliver and Boyd, Edinburgh, 1925.
\newblock URL \url{https://wellcomecollection.org/works/vb9nd67r}.

\bibitem[Grill et~al.(2020)Grill, Strub, Altch{\'e}, Tallec, Richemond,
  Buchatskaya, Doersch, Pires, Guo, Azar, et~al.]{grill2020byol}
Jean-Bastien Grill, Florian Strub, Florent Altch{\'e}, Corentin Tallec,
  Pierre~H. Richemond, Elena Buchatskaya, Carl Doersch, Bernardo~Avila Pires,
  Zhaohan~Daniel Guo, Mohammad~Gheshlaghi Azar, et~al.
\newblock Bootstrap your own latent: A new approach to self-supervised
  learning.
\newblock In \emph{Advances in Neural Information Processing Systems
  (NeurIPS)}, volume~33, pp.\  21271--21284, 2020.
\newblock URL
  \url{https://proceedings.neurips.cc/paper_files/paper/2020/hash/f3ada80d5c4ee70142b17b8192b2958e-Abstract.html}.

\bibitem[Hamilton(1989)]{hamilton1989hmm}
James~D. Hamilton.
\newblock A new approach to the economic analysis of nonstationary time series
  and the business cycle.
\newblock \emph{Econometrica}, 57\penalty0 (2):\penalty0 357--384, 1989.
\newblock \doi{10.2307/1912559}.
\newblock URL \url{https://users.ssc.wisc.edu/~bhansen/718/Hamilton1989.pdf}.

\bibitem[He et~al.(2020)He, Fan, Wu, Xie, and Girshick]{he2020moco}
Kaiming He, Haoqi Fan, Yuxin Wu, Saining Xie, and Ross Girshick.
\newblock Momentum contrast for unsupervised visual representation learning.
\newblock In \emph{Proceedings of the IEEE/CVF Conference on Computer Vision
  and Pattern Recognition (CVPR)}, pp.\  9729--9738, 2020.
\newblock URL \url{https://arxiv.org/abs/1911.05722}.

\bibitem[Holm(1979)]{holm1979simple}
Sture Holm.
\newblock A simple sequentially rejective multiple test procedure.
\newblock \emph{Scandinavian Journal of Statistics}, 6\penalty0 (2):\penalty0
  65--70, 1979.
\newblock URL \url{https://www.jstor.org/stable/4615733}.

\bibitem[Horvath et~al.(2024)Horvath, Issa, and
  Muguruza]{horvath2024clustering}
Blanka Horvath, Zacharia Issa, and Aitor Muguruza.
\newblock Clustering market regimes using the {Wasserstein} distance.
\newblock \emph{Journal of Computational Finance}, 28\penalty0 (1):\penalty0
  1--39, 2024.
\newblock \doi{10.21314/JCF.2024.005}.
\newblock URL \url{https://arxiv.org/abs/2110.11848}.

\bibitem[Hubert \& Arabie(1985)Hubert and Arabie]{hubert1985ari}
Lawrence Hubert and Phipps Arabie.
\newblock Comparing partitions.
\newblock \emph{Journal of Classification}, 2\penalty0 (1):\penalty0 193--218,
  1985.
\newblock \doi{10.1007/BF01908075}.

\bibitem[Killick et~al.(2012)Killick, Fearnhead, and Eckley]{killick2012pelt}
R.~Killick, P.~Fearnhead, and I.~A. Eckley.
\newblock Optimal detection of changepoints with a linear computational cost.
\newblock \emph{Journal of the American Statistical Association}, 107\penalty0
  (500):\penalty0 1590--1598, 2012.
\newblock \doi{10.1080/01621459.2012.737745}.
\newblock URL \url{https://arxiv.org/abs/1101.1438}.

\bibitem[Kingma \& Welling(2014)Kingma and Welling]{kingma2014vae}
Diederik~P. Kingma and Max Welling.
\newblock Auto-encoding variational {B}ayes.
\newblock In \emph{International Conference on Learning Representations
  (ICLR)}, 2014.
\newblock URL \url{https://arxiv.org/abs/1312.6114}.

\bibitem[Kost \& McDermott(2002)Kost and McDermott]{kost2002combined}
James~T. Kost and Michael~P. McDermott.
\newblock Combining dependent {P}-values.
\newblock \emph{Statistics \& Probability Letters}, 60\penalty0 (2):\penalty0
  183--190, 2002.
\newblock \doi{10.1016/S0167-7152(02)00310-3}.
\newblock URL
  \url{https://ideas.repec.org/a/eee/stapro/v60y2002i2p183-190.html}.

\bibitem[K{\"u}nsch(1989)]{kunsch1989jackknife}
Hans~R. K{\"u}nsch.
\newblock The jackknife and the bootstrap for general stationary observations.
\newblock \emph{The Annals of Statistics}, 17\penalty0 (3):\penalty0
  1217--1241, 1989.
\newblock \doi{10.1214/aos/1176347265}.
\newblock URL
  \url{https://projecteuclid.org/journals/annals-of-statistics/volume-17/issue-3/The-Jackknife-and-the-Bootstrap-for-General-Stationary-Observations/10.1214/aos/1176347265.full}.

\bibitem[Lee et~al.(2022)Lee, Aune, Langet, and Eidsvik]{lee2022vnibcreg}
Daesoo Lee, Erlend Aune, Nad{\`e}ge Langet, and Jo~Eidsvik.
\newblock {VNIbCReg}: {VICReg} with neighboring-invariance and
  better-covariance evaluated on non-stationary seismic signal time series.
\newblock \emph{arXiv preprint arXiv:2204.02697}, 2022.
\newblock URL \url{https://arxiv.org/abs/2204.02697}.

\bibitem[Longin \& Solnik(2001)Longin and Solnik]{longin2001extreme}
Fran{\c{c}}ois Longin and Bruno Solnik.
\newblock Extreme correlation of international equity markets.
\newblock \emph{The Journal of Finance}, 56\penalty0 (2):\penalty0 649--676,
  2001.
\newblock \doi{10.1111/0022-1082.00340}.
\newblock URL
  \url{https://longin.fr/Recherche_Publications/Articles_pdf/Longin_Solnik_Extreme_corelation_of_international_equity_market.pdf}.

\bibitem[Ma et~al.(2022)Ma, Ventre, and Polukarov]{ma2022denoised}
Yanqing Ma, Carmine Ventre, and Maria Polukarov.
\newblock Denoised labels for financial time series data via self-supervised
  learning.
\newblock In \emph{Proceedings of the Third ACM International Conference on AI
  in Finance (ICAIF)}, pp.\  471--479, 2022.
\newblock \doi{10.1145/3533271.3561687}.
\newblock URL \url{https://arxiv.org/abs/2112.10139}.

\bibitem[Merchant et~al.(2026)Merchant, Guthrie, Mahns, Balestriero, and
  Levy]{merchant2026market1t}
Humzah Merchant, Alec Guthrie, Simon Mahns, Randall Balestriero, and Bradford
  Levy.
\newblock Towards financial world modeling.
\newblock arXiv preprint arXiv:2610.09048, 2026.
\newblock URL \url{https://arxiv.org/abs/2610.09048}.

\bibitem[Musgrave et~al.(2020)Musgrave, Belongie, and Lim]{musgrave2020reality}
Kevin Musgrave, Serge Belongie, and Ser-Nam Lim.
\newblock A metric learning reality check.
\newblock In \emph{European Conference on Computer Vision (ECCV)}, pp.\
  681--699, 2020.
\newblock URL
  \url{https://www.ecva.net/papers/eccv_2020/papers_ECCV/html/5170_ECCV_2020_paper.php}.

\bibitem[Nystrup et~al.(2020)Nystrup, Lindstr{\"o}m, and
  Madsen]{nystrup2020learning}
Peter Nystrup, Erik Lindstr{\"o}m, and Henrik Madsen.
\newblock Learning hidden {Markov} models with persistent states by penalizing
  jumps.
\newblock \emph{Expert Systems with Applications}, 150:\penalty0 113307, 2020.
\newblock \doi{10.1016/j.eswa.2020.113307}.
\newblock URL
  \url{https://backend.orbit.dtu.dk/ws/files/255194701/Learning_hidden_Markov_models_with_persistent_states_by_penalizing_jumps_ACCEPTED_ESWA.pdf}.

\bibitem[Nystrup et~al.(2021)Nystrup, Kolm, and
  Lindstr{\"o}m]{nystrup2021feature}
Peter Nystrup, Petter~N. Kolm, and Erik Lindstr{\"o}m.
\newblock Feature selection in jump models.
\newblock \emph{Expert Systems with Applications}, 184:\penalty0 115558, 2021.
\newblock \doi{10.1016/j.eswa.2021.115558}.
\newblock URL
  \url{https://backend.orbit.dtu.dk/ws/files/266444218/1_s2.0_S0957417421009647_main.pdf}.

\bibitem[Politis \& Romano(1992)Politis and Romano]{politis1992circular}
Dimitris~N. Politis and Joseph~P. Romano.
\newblock A circular block-resampling procedure for stationary data.
\newblock In Raoul LePage and Lynne Billard (eds.), \emph{Exploring the Limits
  of Bootstrap}, pp.\  263--270. Wiley, New York, 1992.
\newblock URL
  \url{https://statistics.stanford.edu/technical-reports/circular-block-resampling-procedure-stationary-data}.

\bibitem[Rabiner(1989)]{rabiner1989tutorial}
Lawrence~R. Rabiner.
\newblock A tutorial on hidden {M}arkov models and selected applications in
  speech recognition.
\newblock \emph{Proceedings of the IEEE}, 77\penalty0 (2):\penalty0 257--286,
  1989.
\newblock \doi{10.1109/5.18626}.
\newblock URL
  \url{https://www.ee.columbia.edu/~dpwe/e6820/papers/Rabiner89-hmm.pdf}.

\bibitem[Rousseeuw(1987)]{rousseeuw1987silhouette}
Peter~J. Rousseeuw.
\newblock Silhouettes: A graphical aid to the interpretation and validation of
  cluster analysis.
\newblock \emph{Journal of Computational and Applied Mathematics}, 20:\penalty0
  53--65, 1987.
\newblock \doi{10.1016/0377-0427(87)90125-7}.
\newblock URL
  \url{https://wis.kuleuven.be/stat/robust/papers/publications-1987/rousseeuw-silhouettes-jcam-sciencedirectopenarchiv.pdf}.

\bibitem[Sahm(2019)]{sahm2019rule}
Claudia Sahm.
\newblock Direct stimulus payments to individuals.
\newblock In Heather Boushey, Ryan Nunn, and Jay Shambaugh (eds.),
  \emph{Recession Ready: Fiscal Policies to Stabilize the American Economy},
  pp.\  67--92. The Hamilton Project (Brookings Institution) and the Washington
  Center for Equitable Growth, 2019.
\newblock URL
  \url{https://www.hamiltonproject.org/assets/files/Sahm_web_20190506.pdf}.

\bibitem[Satterthwaite(1946)]{satterthwaite1946approximate}
F.~E. Satterthwaite.
\newblock An approximate distribution of estimates of variance components.
\newblock \emph{Biometrics Bulletin}, 2\penalty0 (6):\penalty0 110--114, 1946.
\newblock \doi{10.2307/3002019}.
\newblock URL \url{https://www.jstor.org/stable/3002019}.

\bibitem[Tonekaboni et~al.(2021)Tonekaboni, Eytan, and
  Goldenberg]{tonekaboni2021tnc}
Sana Tonekaboni, Danny Eytan, and Anna Goldenberg.
\newblock Unsupervised representation learning for time series with temporal
  neighborhood coding.
\newblock In \emph{International Conference on Learning Representations
  (ICLR)}, 2021.
\newblock URL \url{https://arxiv.org/abs/2106.00750}.

\bibitem[Trirat et~al.(2026)Trirat, Shin, Kang, Nam, Na, Bae, Kim, Kim, and
  Lee]{trirat2026universal}
Patara Trirat, Yooju Shin, Junhyeok Kang, Youngeun Nam, Jihye Na, Minyoung Bae,
  Joeun Kim, Byunghyun Kim, and Jae-Gil Lee.
\newblock Universal time-series representation learning: A survey.
\newblock \emph{ACM Computing Surveys}, 58\penalty0 (12):\penalty0 1--40, 2026.
\newblock \doi{10.1145/3817600}.
\newblock URL \url{https://arxiv.org/abs/2401.03717}.

\bibitem[Vostrikova(1981)]{vostrikova1981binary}
L.~Yu. Vostrikova.
\newblock Detecting ``disorder'' in multidimensional random processes.
\newblock \emph{Doklady Akademii Nauk SSSR}, 259\penalty0 (2):\penalty0
  270--274, 1981.
\newblock URL \url{https://www.mathnet.ru/eng/dan44582}.

\bibitem[Welch(1947)]{welch1947generalization}
B.~L. Welch.
\newblock The generalization of {S}tudent's problem when several different
  population variances are involved.
\newblock \emph{Biometrika}, 34\penalty0 (1/2):\penalty0 28--35, 1947.
\newblock \doi{10.1093/biomet/34.1-2.28}.
\newblock URL \url{https://pubmed.ncbi.nlm.nih.gov/20287819/}.

\bibitem[Whaley(1993)]{whaley1993vix}
Robert~E. Whaley.
\newblock Derivatives on market volatility: Hedging tools long overdue.
\newblock \emph{Journal of Derivatives}, 1\penalty0 (1):\penalty0 71--84, 1993.
\newblock \doi{10.3905/jod.1993.407868}.
\newblock URL \url{https://www.whaley.info/research-articles/1990-1999}.

\bibitem[Woo et~al.(2022)Woo, Liu, Sahoo, Kumar, and Hoi]{woo2022cost}
Gerald Woo, Chenghao Liu, Doyen Sahoo, Akshat Kumar, and Steven C.~H. Hoi.
\newblock {CoST}: Contrastive learning of disentangled seasonal-trend
  representations for time series forecasting.
\newblock In \emph{International Conference on Learning Representations
  (ICLR)}, 2022.
\newblock URL \url{https://arxiv.org/abs/2202.01575}.

\bibitem[Woo et~al.(2024)Woo, Liu, Kumar, Xiong, Savarese, and
  Sahoo]{woo2024moirai}
Gerald Woo, Chenghao Liu, Akshat Kumar, Caiming Xiong, Silvio Savarese, and
  Doyen Sahoo.
\newblock Unified training of universal time series forecasting transformers.
\newblock In \emph{International Conference on Machine Learning (ICML)}, volume
  235, pp.\  53140--53164, 2024.
\newblock URL \url{https://proceedings.mlr.press/v235/woo24a.html}.

\bibitem[Yue et~al.(2022)Yue, Wang, Duan, Yang, Huang, Tong, and
  Xu]{yue2022ts2vec}
Zhihan Yue, Yujing Wang, Juanyong Duan, Tianmeng Yang, Congrui Huang, Yunhai
  Tong, and Bixiong Xu.
\newblock {TS2Vec}: Towards universal representation of time series.
\newblock In \emph{AAAI Conference on Artificial Intelligence}, volume~36, pp.\
   8980--8987, 2022.
\newblock \doi{10.1609/aaai.v36i8.20881}.
\newblock URL \url{https://arxiv.org/abs/2106.10466}.

\bibitem[Zbontar et~al.(2021)Zbontar, Jing, Misra, LeCun, and
  Deny]{zbontar2021barlow}
Jure Zbontar, Li~Jing, Ishan Misra, Yann LeCun, and St{\'e}phane Deny.
\newblock {Barlow Twins}: Self-supervised learning via redundancy reduction.
\newblock In \emph{International Conference on Machine Learning (ICML)}, pp.\
  12310--12320, 2021.
\newblock URL \url{https://proceedings.mlr.press/v139/zbontar21a.html}.

\bibitem[Zerveas et~al.(2021)Zerveas, Jayaraman, Patel, Bhamidipaty, and
  Eickhoff]{zerveas2021transformer}
George Zerveas, Srideepika Jayaraman, Dhaval Patel, Anuradha Bhamidipaty, and
  Carsten Eickhoff.
\newblock A transformer-based framework for multivariate time series
  representation learning.
\newblock In \emph{Proceedings of the 27th ACM SIGKDD Conference on Knowledge
  Discovery and Data Mining (KDD)}, pp.\  2114--2124, 2021.
\newblock \doi{10.1145/3447548.3467401}.
\newblock URL \url{https://arxiv.org/abs/2010.02803}.

\end{thebibliography}

\clearpage
\appendix
\makeatletter
\setlength{\@fptop}{0pt}
\makeatother

\section{Method Formulations and Configurations}
\label{app:methods}

The empirical comparisons depend on how each objective produces the
representation that is clustered. We first describe the shared
architecture and its objective variants, then the methods that use
different temporal or generative assumptions. The training and
evaluation procedures specify which choices are shared and which
remain method-specific.

Let $z, z' \in \mathbb{R}^{B \times d}$ be the projector outputs, before
normalisation, of a batch of clean anchor windows and of their augmented
positives, $B$ the batch size and $d{=}16$. Define
$\sigma_j(z) = \sqrt{\mathrm{Var}(z_{\cdot j}) + \epsilon}$ with $\epsilon = 10^{-4}$,
$\mathrm{Cov}(z) = (z - \bar z)^{\top}(z - \bar z)/(B-1)$ and
$\mathrm{Off}(z) = \frac{1}{d}\sum_{i \neq j}\mathrm{Cov}(z)_{ij}^2$.
This notation is shared by the joint-embedding objectives below.

\subsection{Joint-Embedding Objectives and Architecture}
\paragraph{Shared backbone.} The six joint-embedding encoders defined
below map the 30-dimensional window through three residual blocks of
width 512, 256 and 128. Each block contains a linear layer, layer
normalisation, GELU and dropout 0.1. A two-layer projector maps 128 to 128
to 16 with layer normalisation and GELU. Their losses act on
the 16-dimensional projector output $z$ before normalisation.
NT-Xent uses its $L_2$ normalisation. Clustering uses $z$ before normalisation for five
of these encoders and its $L_2$-normalised form for SimCLR+VC. Thus the
reported comparison evaluates each loss through the output used by its
training objective. The representation choices differ across methods
as detailed in Section~\ref{sec:limits}.

\subsubsection{Variance and Redundancy Regularisation}
These objectives encourage agreement between related windows while
constraining how information is distributed across embedding dimensions.
The variants distinguish within-view covariance penalties from
cross-view correlation penalties.

\paragraph{VICReg-TN.} This variant applies the
VICReg objective~\citep{bardes2022vicreg} to the projector output, with
positives drawn from the temporal neighbourhood of a window. Its loss is
\begin{equation}
\mathcal{L}_{\mathrm{VICReg\text{-}TN}} = \mu\,\|z - z'\|_F^2/(Bd)
+ \tfrac{\lambda}{d}\sum_j \bigl[\max(0, \gamma - \sigma_j(z)) + \max(0, \gamma - \sigma_j(z'))\bigr]
+ \nu\,\bigl(\mathrm{Off}(z) + \mathrm{Off}(z')\bigr),
\end{equation}
where $\mu{=}1$, $\lambda{=}25$, $\nu{=}1$, $\gamma{=}1$. The three terms
are VICReg's invariance, variance and covariance terms. A related method,
VNIbCReg~\citep{lee2022vnibcreg}, combines a VICReg-style objective with
temporal neighbourhood coding on seismic series.

\paragraph{Barlow Twins.} The redundancy-reduction objective minimises
$\sum_i (1 - C_{ii})^2 + \lambda_{\mathrm{BT}}\sum_{i \neq j} C_{ij}^2$
on the cross-correlation $C$ of batch-standardised views, with
$\lambda_{\mathrm{BT}} = \sci{5}{-3}$~\citep{zbontar2021barlow}.

\paragraph{Regularisation variants.} BT+VH adds the variance term to
the redundancy-reduction objective. It has no within-view covariance
term and retains the off-diagonal cross-correlation penalty.
VICReg-Hard keeps the temporal-neighbourhood objective's invariance
and variance terms, with the invariance at weight 25, and replaces
the covariance penalty with the cross-correlation objective.

\subsubsection{Similarity Learning and Prediction}
The same backbone also supports contrasting matched views and predicting
a slowly updated target representation. Their regularisation and
normalisation choices are retained in the evaluated embeddings.

\paragraph{SimCLR+VC.} The regularised contrastive objective uses NT-Xent
on $L_2$-normalised projections at temperature
$\tau{=}0.07$~\citep{chen2020simclr}. It adds the variance and covariance
terms defined above on $z$ and $z'$, with $\lambda{=}25$, $\nu{=}1$ and $\gamma{=}1$.
The published contrastive objective does not include these regularisers.

\paragraph{BYOL.} The teacher--student objective minimises the mean squared
error between the normalised online
prediction and the stop-gradient target embedding, symmetrised over the two
views. Its predictor has 16, 128 and 16 units with batch normalisation
and ReLU. Target EMA decay is fixed at $0.996$, the published starting
value~\citep{grill2020byol}. Dropout remains active in the target
encoder during training, unlike
the deterministic target branch in the published method.

\subsection{Temporal Objectives}
Temporal information can enter the learning problem through the choice
of neighbouring windows or through an encoder that processes a sequence.
The neighbourhood discriminator below still embeds each window on its
own. The convolutional methods instead use sequence context, which also
affects their evaluation procedure.

\paragraph{Hierarchical contrast.} TS2Vec-style uses a simplified form of
the TS2Vec hierarchical contrastive loss
of~\citet{yue2022ts2vec} on the dilated CNN described under
\emph{Context and decomposition} below. At
each temporal scale, one cross-entropy at temperature 0.5 matches each timestamp
of the first view to the same timestamp of the second. The views come from
timestamp masking at rate 0.5 and jitter 0.01. The published loss also has
same-view negatives, both directions and an instance-wise term. Training here
reads one source sequence, so there are no other sequences to contrast.

\paragraph{Neighbourhood discrimination.} TNC-style uses temporal
neighbourhood coding~\citep{tonekaboni2021tnc} on a
plain MLP without residual connections or a projector. Its hidden layers
have widths 512, 256 and 128, with layer normalisation, GELU and dropout
0.1, followed by a linear map to 16. A discriminator with widths 32, 64
and 1 and GELU separates a window within 15 steps of the
anchor from five windows drawn outside that range, by an unweighted sum of
binary cross-entropies. The published method sets the neighbourhood with a
stationarity test and weights non-neighbours as positive-unlabelled. Here the
neighbourhood is fixed and there is no such weight.

\paragraph{Context and decomposition.} TS-TCC-style and CoST-style use
the same dilated CNN as hierarchical contrast, with width
64, depth 4 and a 16-dimensional output. The TS-TCC-style encoder
follows~\citet{eldele2021tstcc}. Its weak view uses 5\% per-channel
scaling and jitter 0.01. Its strong view uses 20\% scaling, jitter 0.05
and timestamp masking at rate 0.2. Its loss weights equally a cross-view
per-timestamp contrastive term and a contextual term in which a GRU
summary of one view is contrasted with the next timestamp of the other,
both at temperature 0.2. The published method uses a transformer context
model, hence the name. CoST-style follows~\citet{woo2022cost} with simplified
heads: a trend head that pools the backbone output at scales 2, 4 and 8,
and a seasonal head built from the eight largest DFT amplitudes, each
giving eight of the sixteen dimensions and each trained with its own
per-timestamp contrastive loss at temperature 0.5 with equal weights under
timestamp masking at rate 0.5 and jitter 0.01. The published method uses
a mixture of causal convolution experts for the trend and a
frequency-domain loss. All three convolutional sequence methods train on the source's
whole training sequence, one gradient step per epoch for 100 epochs, with
AdamW at learning rate $10^{-3}$, its default weight decay and cosine
decay without warm-up, at twenty seeds in all five evaluation settings.

\paragraph{Published-objective variants.} To examine the effect of
the simplified context and decomposition objectives, the TS-TCC and CoST variants
retain the dilated CNN family and the benchmark's optimisation schedule
while using the published objectives. The context variant forms a weak
view by scaling each timestep by a draw from $N(2, 1.1)$. Its strong view
uses jitter 0.8 and a permutation into at most seven segments. A four-layer,
four-head pre-norm transformer with a class token predicts six future
steps, and the contextual NT-Xent terms use temperature 0.2 and weights 1
and 0.7. Training uses batches of 128 random crops of length 64.

The decomposition variant uses causal-convolution trend experts with kernels from 1
to 128 and a Fourier seasonal layer. It uses a momentum key encoder with decay 0.999,
a queue of 256, temperature 0.07, and the mean of the amplitude and phase
instance terms at weight $\sci{5}{-4}$. It follows the published section
sampling and augmentation, encodes through a causal sliding window of 201
steps, and trains on batches of at most 32. Both variants differ from their
published methods by using a depth-4, width-64 dilated CNN and
the shared AdamW schedule. The TS-TCC transformer width is also 64. Both
variants run at twenty seeds in all five evaluation settings.

\subsection{Generative Reconstruction}
Reconstruction provides a different training signal from matching or
contrasting views. It also introduces a separate choice of latent
representation and checkpoint, specified here for the generative baseline.

The VAE is a variational autoencoder~\citep{kingma2014vae} with a
diagonal Gaussian posterior, encoder and decoder of width 512, 256 and 128
with batch normalisation and ReLU. Its loss is the mean squared reconstruction error
plus the KL divergence at $\beta{=}1$. The 16-dimensional posterior mean is the
embedding. It trains for 100 epochs in mini-batches of 128 with Adam at learning
rate $10^{-3}$ and no schedule. It chooses its epoch
using the lowest unlabelled loss on the source asset's evaluation windows.
This evaluation-based checkpoint rule places the VAE among the
sensitivity analyses, as explained in Section~\ref{sec:limits}.

\subsection{Training and Evaluation Procedures}
Objective differences must be read alongside the training schedule
and the observations available when an embedding is computed. The
following procedures connect the losses above to the partitions scored
by the benchmark.

\paragraph{Augmentations and optimisation.} The six joint-embedding encoders
share the following training procedure. Each step draws 128 anchors uniformly with replacement
and, for each, a positive uniformly from the other windows within 15 steps.
Only the positive is augmented, by two of three operations chosen at random:
Gaussian jitter with standard deviation 0.3, a scale drawn from $[0.7, 1.3]$, and
setting each feature to zero with probability 0.15. An epoch is
$\lfloor n/128 \rfloor$ steps for $n$ training windows. Training runs 100 epochs
with AdamW at learning rate $10^{-3}$ and weight decay $10^{-6}$, 10 warm-up
epochs, cosine decay to 1\% of the rate and gradients clipped at norm 1, and
keeps the epoch of lowest training loss. TNC-style uses 100 epochs of shuffled
mini-batches of 128, AdamW at $10^{-3}$ with its default weight decay and cosine
decay without warm-up. The convolutional sequence methods use the
whole-sequence schedule specified above. Apart from the six
joint-embedding methods and the generative baseline, every encoder
keeps its last epoch.

\paragraph{Embedding at evaluation.} These training schedules produce
representations with different context requirements. The MLP encoders embed each window on its
own. The sequence models read the evaluation block as one sequence
in date order through causal convolutions. A retained embedding can therefore
depend on earlier excluded boundary windows. CoST-style's seasonal head also selects
frequencies over the complete block, so a window can depend on later windows.
These transductive rows are sensitivity analyses rather than members of the
seven-method pointwise family.

\paragraph{Tuning.} The default settings above provide the reference
comparison. The tuning experiment of Section~\ref{sec:tuning} examines
how much that comparison changes within a shared search budget. It searches,
for the six joint-embedding encoders, learning rate, weight decay, batch size,
embedding width, layer widths, dropout, the temporal neighbourhood and the
augmentation family and the method's own loss settings. The search varies
two loss settings for VICReg-TN and one for each of the other methods. Epochs, warm-up, augmentation strengths and the checkpoint rule
stay fixed. TNC-style searches learning rate, embedding width, layer widths, batch
size, neighbourhood and the number of negatives, and the VAE the first four and
$\beta$. The temporal CNN encoders search learning
rate, embedding width, hidden width and depth. Each configuration has two search
seeds. Its secondary criterion is the proxy-label ARI on the same validation
tail, whose FX labels use the earlier mixed-horizon construction.

\subsection{Reference Representations and Clustering}
The references change the source of the representation or infer market
states directly. The supervised encoder learns from source proxy labels,
the frozen forecaster reuses pretraining, and the classical detectors
operate on target features. The configurations below explain how these
outputs enter the label-alignment task and which sources of randomness
remain in each comparison.

\paragraph{Supervised encoder.} The supervised reference trains the shared
MLP backbone with a softmax head using cross-entropy on the source asset's
own label family for
100 epochs at twenty seeds on training windows only, keeping the last epoch.
Its FX source labels use the earlier mixed-horizon construction, whose thresholds are fitted on
the full sample. The
crypto training years hold two of the crypto family's classes, so the
crypto-source head is binary. The 16-dimensional penultimate embedding
enters the transfer protocol unchanged, and its clustering grid
gives both selector columns of Table~\ref{tab:main}.

\paragraph{Pretrained forecaster.} The frozen reference is
Chronos-T5 Small~\citep{ansari2024chronos}.
Each asset's returns are z-scored separately within each window and then
encoded. Hidden states are mean-pooled over time and then over assets. Seed
variance is clustering initialisation variance only.
Table~\ref{tab:main} reports the mixed-horizon comparison.

\paragraph{Clustering.} On the encoder side every configuration in
$G$ is fitted on the embeddings of each training seed with a
fixed clustering seed: $k$-means with twenty initialisations and a
Gaussian mixture with full covariance and ten initialisations, so the
seed variance is training variance. Raw features are standardised with
the target's training-split statistics and clustered with the same two
algorithms at ten initialisations each. Raw draw $s$ uses clustering
seed $s$, so raw has twenty draws of its own, which share no random draw
with the encoder runs. Covariance regularisation is held fixed across
mixture fits. On the raw features some fits fail in
single precision, because the regulariser is smaller than the rounding
error of a near-singular component covariance. Those fits are repeated
in double precision with the same seed and settings. This affects 204 of
5,200 raw mixture fits, with none requiring a larger regulariser.

\paragraph{Classical detectors.} Under the fixed selector, the Gaussian
HMM has 3 states and full covariance. PELT~\citep{killick2012pelt} uses
$\ell_2$ cost, penalty $2\log T$ for a series of length $T$, and minimum
segment 20. Binary segmentation~\citep{vostrikova1981binary} uses
$\ell_2$ cost and two breakpoints, yielding three segments. A segmenter returns segments, not
regime labels. The earlier mixed-horizon evaluation numbered consecutive segments
$0,1,2,\dots$ and took the number modulo three, so a recurring regime
received a new label at every occurrence. The boundary-and-threshold procedure maps
segments to states by $k$-means with $k{=}3$, seed 42 and ten restarts on the
unweighted feature means of the segments, a segment-count-independent
rule, and the earlier convention is kept as a sensitivity analysis.
The statistical jump model~\citep{nystrup2020learning}
uses three states, jump penalty 50 on daily standardised features, and
ten restarts per seed. Like the HMM, it has seed variance and receives
target-level tests.
For label-aware selection, the HMM state grid is $\{2,3,4,5\}$.
The two segmenters use rbf cost. PELT uses minimum segment 5 and penalty
grid $\{3,5,8,12\}$, while binary segmentation uses minimum
segment 5 and breakpoint grid $\{2,\dots,6\}$. Each grid is cycled
over the twenty seeds, so the label-aware spread is hyperparameter
spread rather than seed variance, which is why PELT and MBD sit outside
the Holm families under both selectors. All four detectors ignore the
source asset, so their six directions are three targets counted twice
and their tests combine the three targets.

\subsection{Statistical Details}
\label{app:inference}
The procedures above yield scores indexed by training or clustering
seed and transfer direction. Directions can share a source encoder or a
target-side baseline draw, which motivates accounting for dependence
when their tests are combined. Resampling market windows addresses a
different source of variation, so its construction is specified separately.

\paragraph{Combining seed-level tests.}
Fisher's statistic is referred to a scaled $\chi^2$ distribution using
the Brown/Kost-McDermott correction and a moment-matched
approximation~\citep{satterthwaite1946approximate}:
\begin{align}
\chi^2 &= -2\sum_{i=1}^{K}\log p_i, \label{eq:fisher}\\
\mathrm{E}[\chi^2] &= 2K, \qquad
\mathrm{Var}[\chi^2] = 4K + 2\sum_{i<j} f(\rho_{ij}), \label{eq:brown}\\
f(\rho) &= 3.263\rho + 0.710\rho^2 + 0.027\rho^3, \label{eq:kmpoly}\\
\chi^2/c &\sim \chi^2_{\mathrm{df}_{\mathrm{eff}}}, \quad
c = \frac{\mathrm{Var}[\chi^2]}{2\,\mathrm{E}[\chi^2]}, \quad
\mathrm{df}_{\mathrm{eff}} = \frac{2\,\mathrm{E}[\chi^2]^2}{\mathrm{Var}[\chi^2]}, \label{eq:km}
\end{align}
Here $K$ is the number of directions compared, six on an all-asset block
and two on an equity--FX block. Source-independent methods instead compare
three targets. The quantity $\rho_{ij}$ is the 20-seed correlation across
directions of the per-seed differences between the encoder and its comparator.
It is estimated for direction pairs that share a source encoder or a target
with a common raw draw. Estimates are floored at zero, and pairs without
either shared component are assigned zero correlation. Under this rule $c \ge 1$, so $p_a$ is never below the naive
$p$ in the rejection region. Combined tests use $p_a$ unless explicitly
labelled otherwise.

\paragraph{Moving-block intervals.}
The primary analysis uses circular blocks of $L{=}60$ evaluation windows
and $R{=}5{,}000$ resamples. Each resample applies the same indices to
every seed's embeddings and to the raw features, refits $k{=}3$
$k$-means with ten initialisations, and records the seed-mean encoder
ARI minus raw ARI. The reported interval uses its percentile endpoints.
Raw clustering uses initialisation seed 42 in each resample, whereas the main
tables average twenty raw clustering seeds. The corresponding point
differences differ by at most 0.006, and the intervals condition on
this raw initialisation. An earlier analysis without the boundary
rule used 500 replicates at block lengths 30, 60 and 90. Median
interval widths differed by less than 0.02 across those lengths.
One interval lay wholly above zero, at length 30.

\paragraph{Temporal information.}
Regime episodes are maximal runs of a constant label. We also report
the effective sample size under a first-order autoregressive approximation
\begin{equation}
N_{\mathrm{eff}} = N\,\frac{1-\rho_1}{1+\rho_1}, \label{eq:neff}
\end{equation}
where $N$ is the number of windows and $\rho_1$ is the median lag-1
autocorrelation across feature dimensions. Both summaries describe
dependence within the observed evaluation period.

\section{Calibration of the Seed-Level Test}
\label{app:calibration}
The seed-level statistic of Section~\ref{sec:inference} combines
directions that share source checkpoints or target-side raw draws.
We examine its size and power using the nine encoders reported below,
with per-seed fixed-selector scores from the aligned 2023--24 evaluation. The two checks
below distinguish an empirical row-swap diagnostic from simulation
under the fitted dependence model.

\paragraph{Joint-swap check.} Each side's mean is removed per direction
before encoder and raw rows with seed label $s$ are swapped across all
directions together, over 2{,}000 swaps. Centring prevents a true mean
difference from inducing negative covariance between the swapped groups.
The diagnostic depends on row order because encoder and raw draws are
unmatched and have different variances. Applying one swap pattern to
all directions also imposes stronger coupling than the training design:
different sources reuse seed labels but are separate runs, with
VICReg-TN additionally sharing initial weights.

\paragraph{Parametric null.} We simulate seed effects shared within a source for encoders and
identical baseline draws shared within a target for raw features. The
2{,}000 simulations use the observed variances and within-source
correlations, with equal means on both sides. A coupled scenario adds a
per-seed shock common to all six directions carrying 30\% of the encoder
seed variance. Power is the rejection rate when the encoder mean of one
direction, equity $\to$ FX, is shifted by 0.02, 0.05 or 0.10 ARI over
1{,}000 simulations.

\begin{figure}[H]
\centering
\includegraphics[width=\textwidth]{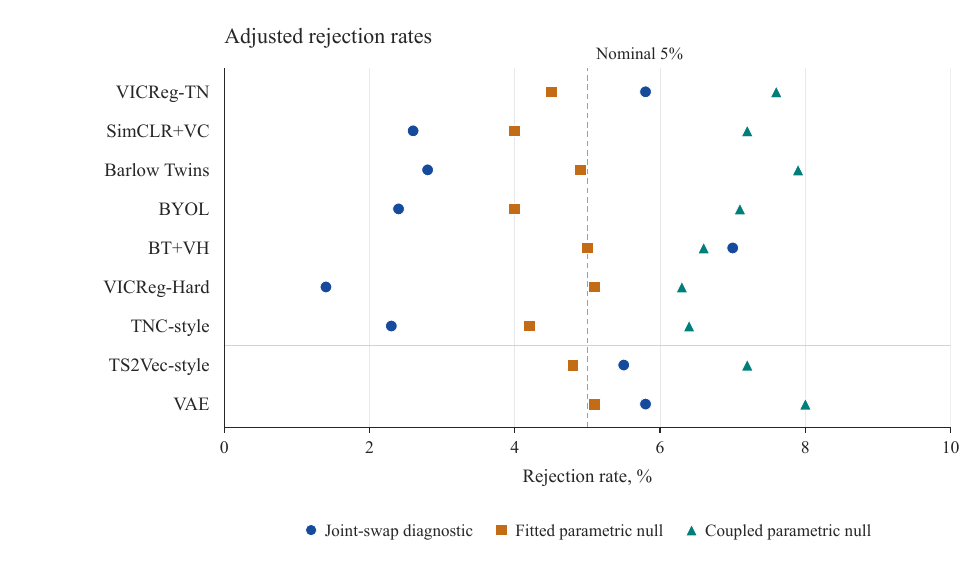}
\caption{Brown-adjusted test calibration in the aligned 2023--24 evaluation
under fixed $k{=}3$ $k$-means. Points give rejection percentages for centred joint
swaps and fitted parametric nulls, with the coupled null adding a shared
seed shock. The dashed line marks the nominal 5\% rate. The divider
separates primary and sensitivity rows. Naive-test rates and power appear
in Table~\ref{tab:calibration}.}
\label{fig:calibration_adjusted}
\end{figure}

Figure~\ref{fig:calibration_adjusted} shows approximately nominal rejection rates
under the fitted dependence model and mildly higher rates when a shared
shock spans all directions. Power increases with the effect in the
shifted direction. This calibration concerns a favourable direction
under the simulated dependence structure. Average effects and temporal
stability require the separate analyses in the main text.
\subsection{Pairing and Covariance Sensitivity}
\label{app:pairing}
The calibration fixes a model of shared seed variation. We also check
whether the observed conclusions depend on matching encoder and raw
rows that do not share a random draw, or on how cross-direction
covariance is estimated.

Of the 242 method--direction cells in the audit over five evaluation
settings, 220 have variance on both sides. The excluded 22 are the equity
$\to$ crypto and FX $\to$ crypto cells for all eleven reported methods in
2021--22, where raw $k{=}3$ clustering is constant across its twenty seeds in
those two directions. Across the remaining cells, the encoder--raw seed
correlation has mean 0.03 and the spread expected under independence.

In the eleven-method sensitivity analysis, pairing rows by seed label
changes no Holm or direction-level decision under the fixed and
label-aware selectors. It changes one Holm decision and one
direction-level decision under $k{=}3$ GMM. In the earlier
mixed-horizon evaluation shown in Table~\ref{tab:main}, arbitrary pairing
removes BT+VH under the fixed selector and BYOL under label-aware
selection. Over 1{,}000 random re-matchings, those comparisons retain
Holm significance in 61\% and 43\% of cases, respectively. The
sensitivity to unmatched row order supports reporting the unpaired
tests used in the main analysis.

The plug-in $\rho_{ij}$ also takes encoder and raw rows in seed order.
Estimating it from the design instead---encoder covariance for a shared
source or target and raw covariance for a shared target---matches no
encoder row to a raw row. This change affects none of the 187 Holm
decisions against raw: 165 cover the five evaluation settings under three
selectors and 22 cover the earlier evaluation. It also affects none of the
16 VICReg-TN head-to-heads in the mixed-horizon setting. The reported $p_a$
moves by a factor of at most 1.9 between $10^{-6}$ and 0.1 and, down to the
$10^{-15}$ display floor, by at most 10.2. Values at that floor should be
read only to this precision.

\section{Wins Carried by One Direction}
\label{app:saturation}

Calibration describes rejection behaviour, but a well-calibrated
combined test can still be driven by an isolated improvement. A small
direction-level $p$-value can dominate the Fisher combination. In the
earlier mixed-horizon evaluation, the fixed selector produces isolated
encoder direction wins. Only BT+VH survives the across-encoder Holm
correction.
Table~\ref{tab:saturated} shows how the label-aware selector concentrates
large seed-level statistics in the equity $\to$ FX direction. These $t$
statistics measure initialisation variance within one period. The
moving-block intervals for the fixed-selector direction wins cross zero
as shown in Table~\ref{tab:primary_dir}.

\begin{table}[h]
\centering
\small
\caption{Equity $\to$ FX comparison with raw features under label-aware
selection in the earlier mixed-horizon evaluation. $\Delta$ is the
encoder-minus-raw mean ARI. $t$ and $p$ are the Welch statistic and its
one-sided upper-tail probability.}
\label{tab:saturated}
\begin{tabular}{lccccc}
\toprule
Encoder & Mean ARI & Baseline & $\Delta$ & $t$ & One-sided $p$ \\
\midrule
Barlow Twins & 0.375 & 0.274 & $+0.100$ & 18.85 & $<10^{-15}$ \\
BYOL & 0.303 & 0.274 & $+0.029$ & 7.07 & $\sci{4.3}{-8}$ \\
BT+VH & 0.378 & 0.274 & $+0.104$ & 19.15 & $<10^{-15}$ \\
VAE & 0.330 & 0.274 & $+0.056$ & 6.10 & $\sci{1.0}{-6}$ \\
\bottomrule
\end{tabular}
\end{table}

\section{Sensitivity to the Equity Universe}
\label{app:universe}

The seed-level analyses condition on a chosen set of assets. This
comparison changes that set while retaining the evaluation dates,
labels and trained representations, so that variation in the input
universe can be examined separately.

The primary equity panel is determined by a coverage tie-break and is not a
size-ranked sample. We therefore rebuild the equity features for two
alternatives: the fifty largest complete-history names in the early-2025
capitalisation list and ten random fifty-name panels from the same eligible
set. The target-side equity directions are rescored on the same dates and
VIX labels, without retraining the encoders.

Table~\ref{tab:universe} shows that the positive primary mean effects
shrink on the size-ranked panel and generally remain positive across
the random panels. Individual equity-direction effects are less stable
and often change sign. These panels describe sensitivity to universe
selection. Their overlap, reused checkpoints and use of later coverage
or capitalisation information limit a sampling interpretation.

\begin{table}[t]
\caption{Equity-universe sensitivity under the primary evaluation's
fixed selector. Margin is the six-direction mean ARI difference from
raw, and Wins counts seed-level direction wins. Primary and Size denote
the original and size-ranked panels. Draws gives the range across ten
random panels, and Above counts panels with positive margins.
Only equity-target directions are rescored.}
\label{tab:universe}
\centering
\small
\setlength{\tabcolsep}{5pt}
\begin{tabular}{@{}lccccccc@{}}
\toprule
 & \multicolumn{4}{c}{Margin over baseline} & \multicolumn{3}{c}{Wins / 6} \\
\cmidrule(lr){2-5}\cmidrule(l){6-8}
Method & Primary & Size & Draws & Above & Primary & Size & Draws \\
\midrule
VICReg-TN & $+0.022$ & $+0.010$ & $+0.010$ to $+0.029$ & 10 & 4 & 2 & 2 to 4 \\
SimCLR+VC & $+0.018$ & $+0.009$ & $+0.008$ to $+0.023$ & 10 & 4 & 4 & 4 to 5 \\
Barlow Twins & $-0.024$ & $-0.027$ & $-0.030$ to $-0.014$ & 0 & 2 & 2 & 2 to 4 \\
BYOL & $-0.067$ & $-0.069$ & $-0.076$ to $-0.062$ & 0 & 1 & 0 & 0 to 2 \\
BT+VH & $-0.025$ & $-0.028$ & $-0.030$ to $-0.015$ & 0 & 2 & 2 & 2 to 4 \\
VICReg-Hard & $-0.066$ & $-0.067$ & $-0.074$ to $-0.056$ & 0 & 0 & 0 & 0 to 2 \\
TS2Vec-style & $+0.012$ & $+0.010$ & $-0.002$ to $+0.018$ & 8 & 4 & 4 & 3 to 4 \\
TNC-style & $-0.007$ & $-0.015$ & $-0.018$ to $-0.001$ & 0 & 1 & 1 & 1 to 2 \\
VAE & $-0.019$ & $-0.027$ & $-0.028$ to $-0.016$ & 0 & 2 & 1 & 1 to 2 \\
TS-TCC-style & $-0.047$ & $-0.059$ & $-0.063$ to $-0.040$ & 0 & 0 & 0 & 0 to 2 \\
CoST-style & $-0.037$ & $-0.052$ & $-0.052$ to $-0.035$ & 0 & 1 & 0 & 0 to 2
 \\
\bottomrule
\end{tabular}
\end{table}
\section{Selection and Encoder Comparisons}
\label{app:selector_analysis}

Universe selection changes the input data. Here we keep the recorded
representations fixed and examine how selecting a partition changes
the comparison. The earlier mixed-horizon evaluation also illustrates
why improving on another encoder and improving on raw features are
distinct claims.

\begin{table}[t]
\caption{Method comparisons with raw features in the earlier
mixed-horizon evaluation. $p_a$ is Brown-adjusted Fisher $p$.
Mean ARI, Wins and Boot.\ use fixed selection. Wins counts seed-level
direction rejections, and Boot.\ counts 95\% block-bootstrap intervals
wholly above zero. The final column uses a correlation-only raw baseline.
$^{\dagger}$ marks Holm significance over the eleven encoder rows,
adding HMM and jump for the first two $p$-value columns. Chronos and the detectors
aggregate three targets rather than six directions. $^{\ddagger}$ counts
positive point differences for methods without seed variance. PELT and
MBD use segments as labels.}
\label{tab:main}
\centering
\scriptsize
\setlength{\tabcolsep}{3.5pt}
\begin{tabular}{@{}lcccccc@{}}
\toprule
Method vs baseline & Label-aware $p_a$ & Fixed $p_a$ & Wins / 6 & Mean ARI & Boot. & vs correlation-only baseline, $p_a$ \\
\midrule
\multicolumn{7}{@{}l}{\emph{SSL encoders}} \\
VICReg-TN & $\sci{2.6}{-6}$\,$^{\dagger}$ & $0.267$ & 1 & 0.129 & 0 & $\sci{1.6}{-10}$\,$^{\dagger}$ \\
SimCLR+VC & $0.049$ & $0.478$ & 1 & 0.117 & 0 & $\sci{4.5}{-5}$\,$^{\dagger}$ \\
Barlow Twins & $\sci{2.5}{-12}$\,$^{\dagger}$ & $0.131$ & 1 & 0.122 & 0 & $<10^{-15}$\,$^{\dagger}$ \\
BYOL & $0.002$\,$^{\dagger}$ & $1.000$ & 0 & 0.083 & 0 & $\sci{7.1}{-10}$\,$^{\dagger}$ \\
BT+VH & $\sci{7.1}{-13}$\,$^{\dagger}$ & $\sci{4.2}{-4}$\,$^{\dagger}$ & 1 & 0.127 & 0 & $<10^{-15}$\,$^{\dagger}$ \\
VICReg-Hard & $1.000$ & $1.000$ & 0 & 0.054 & 0 & $<10^{-15}$\,$^{\dagger}$ \\
TS2Vec-style & $0.005$\,$^{\dagger}$ & $0.961$ & 0 & 0.112 & 0 & $0.036$ \\
TNC-style & $0.997$ & $1.000$ & 0 & 0.101 & 0 & $\sci{1.3}{-9}$\,$^{\dagger}$ \\
VAE & $\sci{6.7}{-4}$\,$^{\dagger}$ & $1.000$ & 0 & 0.114 & 0 & $<10^{-15}$\,$^{\dagger}$ \\
TS-TCC-style & $1.000$ & $1.000$ & 0 & 0.071 & n/a & $0.999$ \\
CoST-style & $0.289$ & $0.981$ & 0 & 0.100 & n/a & $0.018$ \\
\midrule
\multicolumn{7}{@{}l}{\emph{Supervised and foundation-model baselines}} \\
Supervised CE & $0.146$ & $1.000$ & 0 & 0.048 & n/a & n/a \\
Chronos, zero-shot & $1.000$ & $1.000$ & 0 / 3 & 0.019 & n/a & n/a \\
\midrule
\multicolumn{7}{@{}l}{\emph{Classical detectors}} \\
HMM & $1.000$ & $1.000$ & 0 / 3 & 0.099 & n/a & n/a \\
Jump model & $\sci{3.4}{-12}$\,$^{\dagger}$ & $\sci{1.5}{-4}$\,$^{\dagger}$ & 1 / 3 & 0.115 & n/a & n/a \\
PELT & n/a & n/a & 0 / 3$^{\ddagger}$ & 0.049 & 0 & n/a \\
MBD / BinSeg & n/a & n/a & 2 / 3$^{\ddagger}$ & 0.258 & 0 & n/a \\
\bottomrule
\end{tabular}
\end{table}

Table~\ref{tab:main} compares methods with the raw-feature baseline in
the earlier mixed-horizon evaluation. Table~\ref{tab:encoder_comparison}
then examines VICReg-TN against the other encoders under label-aware
selection. All eight head-to-head comparisons survive Holm correction,
as they also do under the fixed selector. Under the latter, however,
VICReg-TN does not establish an improvement over raw. Relative
performance among encoders therefore need not imply progress over
the baseline.

A hierarchical bootstrap resamples directions and then seeds over
5{,}000 replicates. Table~\ref{tab:encoder_comparison} reports the
share retaining Holm significance. This resampling holds the market
observations fixed. Table~\ref{tab:ci} gives the corresponding
per-direction means, while
Table~\ref{tab:primary_dir} reports fixed-selector differences from
raw with moving-block intervals over evaluation windows.

\begin{table}[h]
\centering
\small
\caption{VICReg-TN comparisons under label-aware selection in the earlier
mixed-horizon evaluation. The $p$-values combine one-sided Welch tests
across six directions using Fisher's method, with and without Brown
adjustment. Bootstrap survival is the fraction of hierarchical resamples
of directions and seeds retaining Holm significance across eight pairs.}
\label{tab:encoder_comparison}
\begin{tabular}{lccc}
\toprule
VICReg-TN vs & Naive $p$ & Brown-adjusted $p$ & Bootstrap survival \\
\midrule
SimCLR+VC & $\sci{2.2}{-9}$ & $\sci{2.5}{-8}$ & 0.895 \\
Barlow Twins & $\sci{1.5}{-8}$ & $\sci{2.9}{-8}$ & 0.886 \\
BYOL & $\sci{8.0}{-6}$ & $\sci{1.7}{-5}$ & 0.887 \\
BT+VH & $\sci{7.4}{-7}$ & $\sci{1.1}{-6}$ & 0.852 \\
VICReg-Hard & $<10^{-15}$ & $<10^{-15}$ & 0.987 \\
TS2Vec-style & $\sci{1.7}{-4}$ & $\sci{2.3}{-4}$ & 0.884 \\
TNC-style & $<10^{-15}$ & $\sci{9.9}{-15}$ & 0.906 \\
VAE & $<10^{-15}$ & $<10^{-15}$ & 0.913 \\
\bottomrule
\end{tabular}
\end{table}

\begin{table}[h]
\centering
\footnotesize
\caption{Earlier mixed-horizon evaluation under label-aware selection.
Cells show mean test ARI.
C, E and F denote crypto, equity and FX.}
\label{tab:ci}
\begin{tabular}{@{}lccc@{}}
\toprule
Method & C$\to$E & C$\to$F & E$\to$C \\
\midrule
VICReg-TN & $+0.122$ & $+0.235$ & $+0.126$ \\
SimCLR+VC & $+0.123$ & $+0.210$ & $+0.079$ \\
Barlow Twins & $+0.113$ & $+0.240$ & $+0.092$ \\
BYOL & $+0.094$ & $+0.271$ & $+0.050$ \\
BT+VH & $+0.112$ & $+0.240$ & $+0.094$ \\
VICReg-Hard & $+0.124$ & $+0.219$ & $+0.043$ \\
TS2Vec-style & $+0.096$ & $+0.230$ & $+0.112$ \\
TNC-style & $+0.115$ & $+0.250$ & $+0.044$ \\
VAE & $+0.116$ & $+0.289$ & $+0.058$ \\
\bottomrule
\end{tabular}

\medskip

\begin{tabular}{@{}lccc@{}}
\toprule
Method & E$\to$F & F$\to$C & F$\to$E \\
\midrule
VICReg-TN & $+0.251$ & $+0.113$ & $+0.095$ \\
SimCLR+VC & $+0.256$ & $+0.115$ & $+0.091$ \\
Barlow Twins & $+0.375$ & $+0.076$ & $+0.114$ \\
BYOL & $+0.303$ & $+0.080$ & $+0.098$ \\
BT+VH & $+0.378$ & $+0.086$ & $+0.113$ \\
VICReg-Hard & $+0.261$ & $-0.045$ & $+0.080$ \\
TS2Vec-style & $+0.217$ & $+0.143$ & $+0.095$ \\
TNC-style & $+0.278$ & $+0.072$ & $+0.095$ \\
VAE & $+0.330$ & $+0.004$ & $+0.114$ \\
\bottomrule
\end{tabular}

\end{table}

\begin{table}[!t]
\centering
\footnotesize
\setlength{\tabcolsep}{3pt}
\caption{Earlier mixed-horizon evaluation under fixed $k{=}3$ $k$-means. Cells show mean test ARI, with 95\% moving-block intervals
for the difference from raw features in brackets.
C, E and F denote crypto, equity and FX.}
\label{tab:primary_dir}
\begin{tabular}{@{}lccc@{}}
\toprule
Encoder & C$\to$E & C$\to$F & E$\to$C \\
\midrule
VICReg-TN & 0.101 [-0.10, +0.07] & 0.175 [-0.18, +0.03] & 0.121 [-0.09, +0.22] \\
SimCLR+VC & 0.082 [-0.10, +0.07] & 0.151 [-0.20, +0.04] & 0.071 [-0.12, +0.18] \\
Barlow Twins & 0.098 [-0.10, +0.03] & 0.162 [-0.23, +0.02] & 0.082 [-0.17, +0.26] \\
BYOL & 0.093 [-0.11, +0.02] & 0.194 [-0.21, +0.04] & -0.036 [-0.28, +0.10] \\
BT+VH & 0.097 [-0.10, +0.03] & 0.163 [-0.23, +0.02] & 0.092 [-0.15, +0.26] \\
VICReg-Hard & 0.075 [-0.20, +0.08] & 0.069 [-0.35, -0.04] & -0.043 [-0.32, +0.17] \\
TS2Vec-style & 0.074 [-0.12, +0.05] & 0.169 [-0.21, +0.03] & 0.095 [-0.05, +0.24] \\
TNC-style & 0.096 [-0.11, +0.06] & 0.225 [-0.18, +0.08] & -0.051 [-0.24, +0.13] \\
VAE & 0.112 [-0.07, +0.03] & 0.252 [-0.13, +0.09] & -0.031 [-0.26, +0.05] \\
\midrule
Baseline & 0.132 & 0.272 & 0.093 \\
\bottomrule
\end{tabular}

\vspace{4pt}

\begin{tabular}{@{}lccc@{}}
\toprule
Encoder & E$\to$F & F$\to$C & F$\to$E \\
\midrule
VICReg-TN & 0.197 [-0.17, +0.03] & 0.092 [-0.08, +0.20] & 0.089 [-0.10, +0.01] \\
SimCLR+VC & 0.201 [-0.15, +0.07] & 0.109 [-0.06, +0.24] & 0.088 [-0.10, +0.02] \\
Barlow Twins & 0.292 [-0.07, +0.20] & -0.009 [-0.24, +0.14] & 0.109 [-0.11, +0.02] \\
BYOL & 0.220 [-0.18, +0.06] & -0.062 [-0.26, +0.11] & 0.090 [-0.11, +0.02] \\
BT+VH & 0.310 [-0.06, +0.24] & -0.010 [-0.24, +0.15] & 0.108 [-0.11, +0.02] \\
VICReg-Hard & 0.235 [-0.16, +0.08] & -0.083 [-0.35, -0.00] & 0.071 [-0.16, +0.04] \\
TS2Vec-style & 0.160 [-0.22, +0.03] & 0.102 [-0.04, +0.25] & 0.071 [-0.13, +0.02] \\
TNC-style & 0.230 [-0.13, +0.10] & 0.036 [-0.16, +0.18] & 0.070 [-0.14, +0.00] \\
VAE & 0.266 [-0.12, +0.12] & -0.030 [-0.32, +0.07] & 0.112 [-0.11, +0.04] \\
\midrule
Baseline & 0.272 & 0.093 & 0.132 \\
\bottomrule
\end{tabular}
\end{table}

\section{Published-Objective Variants of TS-TCC and CoST}
\label{app:faithful}

\begin{figure}[H]
\centering
\includegraphics[width=\textwidth]{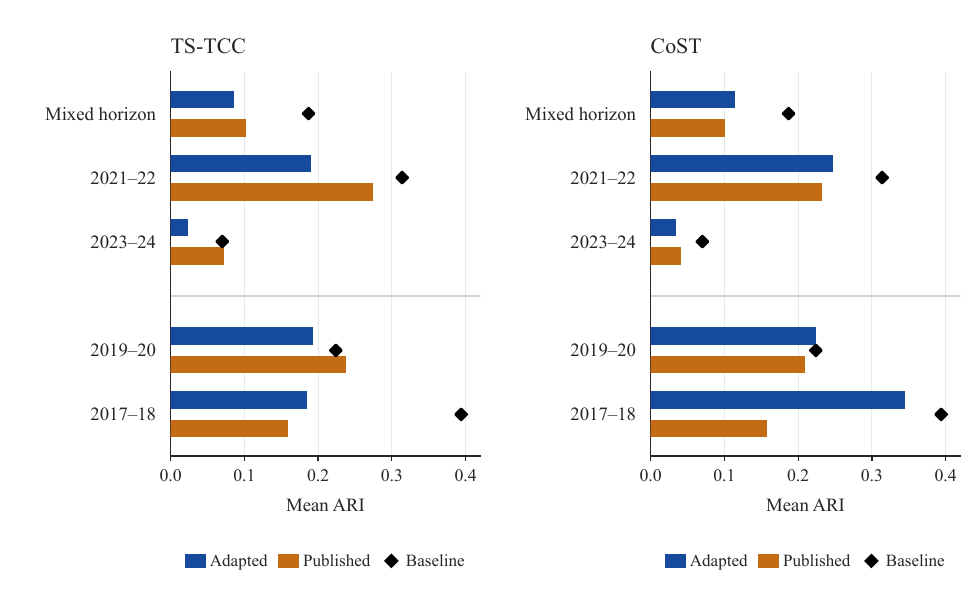}
\caption{Adapted and published-objective variants under fixed $k{=}3$ $k$-means
with the boundary-and-threshold rules. Bars show mean ARI and diamonds
show the raw baseline. The upper three settings average six transfer
directions and the lower two average two equity--FX directions. Full
values, direction wins and corrected-test results are in
Table~\ref{tab:faithful}.}
\label{fig:objective_variants}
\end{figure}

The temporal methods described in Appendix~\ref{app:methods} simplify
parts of the published objectives. To assess the contribution of those
adaptations, Figure~\ref{fig:objective_variants} compares them with the
published-objective variants described above. Those variants retain
the backbone family while restoring objective components and their
associated training choices. The context variant improves on its
adaptation in most settings, with a reversal in the earliest equity--FX
block, while the decomposition comparison varies across
settings. Neither variant has a moving-block
interval wholly above zero, and adding both
variants to the multiplicity family changes no decision for the other
methods. Much of the context method's deficit is therefore sensitive
to these adaptations, while the available intervals still do not
establish a stable advantage over raw features.

\section{Control Representations}
\label{app:controls}

The objective variants change the learned embedding geometry. Controls
help assess how much label alignment can also arise from projection
and architecture without the corresponding training objective.

The controls use the same thirty inputs as the learned methods but no
predictive, supervised or self-supervised objective. They include a
16-component PCA fitted on the source training windows and untrained copies
of the MLP and the context and decomposition backbones. Table~\ref{tab:controls}
shows that PCA and untrained backbones can reach or exceed several trained
methods in the earlier mixed-horizon evaluation. Architecture,
projection and clustering geometry therefore contribute to the measured
score alongside the training objective.

\begin{table}[h]
\centering
\small
\caption{Control representations versus raw features in the earlier
mixed-horizon evaluation, with and without the boundary-and-threshold
rules. Mean ARI averages six transfer directions, Wins counts seed-level
direction wins, and $p_a$ is Brown-adjusted Fisher $p$.}
\label{tab:controls}
\begin{tabular}{lcccccc}
\toprule
& \multicolumn{3}{c}{Label-aware} & \multicolumn{3}{c}{Fixed selector} \\
\cmidrule(lr){2-4}\cmidrule(lr){5-7}
Control & $p_a$ & Wins & Mean ARI & $p_a$ & Wins & Mean ARI \\
\midrule
\multicolumn{7}{@{}l}{\emph{Earlier evaluation}} \\
PCA, 16 components & $0.993$ & 0 & 0.156 & $0.985$ & 0 & 0.153 \\
Untrained encoder & $1.000$ & 0 & 0.115 & $1.000$ & 0 & 0.086 \\
Untrained TS-TCC-style backbone & $1.000$ & 0 & 0.116 & $1.000$ & 0 & 0.093 \\
Untrained CoST-style backbone & $1.000$ & 0 & 0.127 & $1.000$ & 0 & 0.101 \\
Baseline & n/a & n/a & 0.166 & n/a & n/a & 0.166
 \\
\midrule
\multicolumn{7}{@{}l}{\emph{With boundary-and-threshold rules}} \\
PCA, 16 components & $0.996$ & 0 & 0.186 & $0.990$ & 0 & 0.175 \\
Untrained encoder & $0.295$ & 2 & 0.132 & $0.988$ & 0 & 0.100 \\
Untrained TS-TCC-style backbone & $0.261$ & 2 & 0.132 & $0.943$ & 0 & 0.103 \\
Untrained CoST-style backbone & $0.402$ & 2 & 0.146 & $1.000$ & 0 & 0.114 \\
Baseline & n/a & n/a & 0.190 & n/a & n/a & 0.187
 \\
\bottomrule
\end{tabular}
\end{table}
\section{Classical Detectors and Moving-Block Intervals}
\label{app:classical}
The controls retain the representation-then-clustering structure of the
benchmark. Classical detectors offer a complementary comparison by
inferring states or segments directly from the target features. Their
intervals allow favourable point estimates to be assessed against
variation in the observed market sequence.

Table~\ref{tab:classical} scores the four classical detectors of
Section~\ref{sec:methods} on the target's raw features in the mixed-horizon
setting and the primary 2023--24 evaluation with the
boundary-and-threshold rules. For comparison, it also reports the earlier
mixed-horizon evaluation without the boundary rule but with the
segment-state mapping alone. The HMM and the jump model have seed
variance from EM and coordinate-descent initialisation and receive a
one-sided Welch test per target combined over the three targets. PELT and
MBD are deterministic under their fixed configuration and are reported as
point estimates minus the baseline seed mean.

\begin{table}[H]
\caption{Classical detectors under fixed selection across evaluation
settings. Each cell gives seed-mean target ARI followed by its difference
from raw, separated by a slash. $p_a$ is the Brown-adjusted Fisher combination over the three
targets and is omitted for deterministic detectors.}
\label{tab:classical}
\centering
\footnotesize
\setlength{\tabcolsep}{3pt}
\begin{tabular}{@{}llcccc@{}}
\toprule
Detector & Block & Crypto & Equity & FX & $p_a$ \\
\midrule
HMM & Mixed horizon, boundary rules & 0.010 / $-0.023$ & 0.095 / $-0.049$ & 0.246 / $-0.138$ & $1.000$ \\
HMM & 2023--24 & 0.010 / $-0.023$ & 0.025 / $-0.006$ & 0.096 / $-0.050$ & $1.000$ \\
HMM & Earlier mixed horizon & 0.022 / $-0.071$ & 0.099 / $-0.033$ & 0.176 / $-0.096$ & $1.000$ \\
PELT & Mixed horizon, boundary rules & 0.206 / $+0.172$ & 0.149 / $+0.005$ & 0.349 / $-0.036$ & n/a \\
PELT & 2023--24 & 0.206 / $+0.172$ & 0.114 / $+0.084$ & 0.134 / $-0.012$ & n/a \\
PELT & Earlier mixed horizon & $-0.022$ / $-0.115$ & 0.137 / $+0.005$ & 0.304 / $+0.032$ & n/a \\
BinSeg & Mixed horizon, boundary rules & 0.055 / $+0.021$ & 0.172 / $+0.029$ & 0.213 / $-0.171$ & n/a \\
BinSeg & 2023--24 & 0.055 / $+0.021$ & 0.128 / $+0.097$ & 0.098 / $-0.049$ & n/a \\
BinSeg & Earlier mixed horizon & 0.433 / $+0.340$ & 0.174 / $+0.042$ & 0.167 / $-0.105$ & n/a \\
Jump model & Mixed horizon, boundary rules & 0.009 / $-0.024$ & 0.151 / $+0.007$ & 0.312 / $-0.072$ & $<10^{-15}$ \\
Jump model & 2023--24 & 0.009 / $-0.024$ & 0.096 / $+0.065$ & 0.085 / $-0.062$ & $\sci{2.4}{-4}$ \\
Jump model & Earlier mixed horizon & $-0.015$ / $-0.108$ & 0.147 / $+0.014$ & 0.214 / $-0.058$ & $\sci{1.5}{-4}$
 \\
\bottomrule
\end{tabular}
\end{table}

\begin{table}[H]
\caption{Paired moving-block uncertainty for seed-mean ARI differences
from raw under the boundary-and-threshold rules. Above zero counts
95\% intervals with positive lower bounds. Half-width is the median
interval half-width. Denominators count transfer directions for encoders
and target assets for source-independent detectors.}
\label{tab:bootstrap}
\centering
\small
\setlength{\tabcolsep}{5pt}
\begin{tabular}{@{}lcccc@{}}
\toprule
 & \multicolumn{2}{c}{Aligned 2023--24 block} & \multicolumn{2}{c}{Mixed-horizon setting} \\
\cmidrule(lr){2-3}\cmidrule(lr){4-5}
Encoder & above zero & half-width & above zero & half-width \\
\midrule
VICReg-TN & 0 / 6 & 0.103 & 0 / 6 & 0.132 \\
SimCLR+VC & 0 / 6 & 0.108 & 0 / 6 & 0.140 \\
Barlow Twins & 0 / 6 & 0.146 & 0 / 6 & 0.159 \\
BYOL & 0 / 6 & 0.185 & 0 / 6 & 0.149 \\
BT+VH & 0 / 6 & 0.143 & 0 / 6 & 0.166 \\
VICReg-Hard & 0 / 6 & 0.191 & 0 / 6 & 0.158 \\
TS2Vec-style & 0 / 6 & 0.149 & 0 / 6 & 0.133 \\
TNC-style & 0 / 6 & 0.141 & 0 / 6 & 0.157 \\
VAE & 0 / 6 & 0.125 & 0 / 6 & 0.134 \\
TS-TCC-style & 0 / 6 & 0.140 & 0 / 6 & 0.137 \\
CoST-style & 0 / 6 & 0.132 & 0 / 6 & 0.124 \\
\midrule
PELT & 0 / 3 & 0.127 & 0 / 3 & 0.125 \\
BinSeg & 0 / 3 & 0.181 & 0 / 3 & 0.212
 \\
\bottomrule
\end{tabular}
\end{table}

\section{The Exogenous Sahm Family}
\label{app:sahm}

Changing the method leaves the source of the evaluation labels fixed.
The following analysis instead changes the label source to examine
agreement with a macroeconomic indicator that is not constructed from
target prices.

The Sahm labels are the only scored family whose rule and data source do not
use prices from the target asset class. Each monthly value is applied from
the fifteenth day of the following month, after the employment report used
to compute it has been published. The analysis uses the equity target and
the fixed selector and reuses the earlier mixed-horizon checkpoints.
The strict transfer comparison includes only the crypto
and FX sources. A separate sensitivity analysis adds the in-domain equity
source.

Table~\ref{tab:sahm_vsraw} shows that adding the in-domain source
increases the number of significant comparisons. Mean ARI remains
close to zero for every method, including raw, under either source set.
The small $p$-values therefore separate partitions with weak agreement
with the Sahm labels. Keeping the source sets distinct makes this
difference between statistical rejection and regime alignment visible.

\begin{table}[h]
\centering
\footnotesize
\setlength{\tabcolsep}{3.5pt}
\caption{Sahm-label agreement for the equity target in the earlier
mixed-horizon evaluation under fixed selection. Sources are crypto and
FX, with equity added in the second group. Means average sources.
Wins counts seed-level wins over raw, and $p_a$ is Brown-adjusted Fisher
$p$. $^{\dagger}$ marks Holm significance across nine methods per source set.}
\label{tab:sahm_vsraw}
\begin{tabular}{@{}lccc ccc@{}}
\toprule
& \multicolumn{3}{c}{Two transfer sources} & \multicolumn{3}{c}{Including equity source} \\
\cmidrule(lr){2-4}\cmidrule(lr){5-7}
Method & Mean ARI & Wins & $p_a$ & Mean ARI & Wins & $p_a$ \\
\midrule
VICReg-TN & 0.006 & 2 / 2 & $\sci{1.4}{-12}$\,$^{\dagger}$ & 0.001 & 3 / 3 & $<10^{-15}$\,$^{\dagger}$ \\
SimCLR+VC & 0.028 & 2 / 2 & $<10^{-15}$\,$^{\dagger}$ & 0.019 & 3 / 3 & $<10^{-15}$\,$^{\dagger}$ \\
Barlow Twins & $-0.025$ & 0 / 2 & 0.902 & $-0.021$ & 1 / 3 & $\sci{1.9}{-6}$\,$^{\dagger}$ \\
BYOL & 0.009 & 2 / 2 & $\sci{1.3}{-10}$\,$^{\dagger}$ & 0.003 & 3 / 3 & $\sci{3.0}{-10}$\,$^{\dagger}$ \\
BT+VH & $-0.028$ & 0 / 2 & 0.946 & $-0.022$ & 1 / 3 & $\sci{3.9}{-7}$\,$^{\dagger}$ \\
VICReg-Hard & $-0.020$ & 1 / 2 & $\sci{4.1}{-4}$\,$^{\dagger}$ & $-0.017$ & 2 / 3 & $\sci{3.1}{-15}$\,$^{\dagger}$ \\
TS2Vec-style & 0.022 & 2 / 2 & $\sci{3.5}{-8}$\,$^{\dagger}$ & 0.010 & 2 / 3 & $\sci{8.5}{-8}$\,$^{\dagger}$ \\
TNC-style & $-0.041$ & 0 / 2 & 1.000 & $-0.024$ & 1 / 3 & $\sci{8.2}{-11}$\,$^{\dagger}$ \\
VAE & $-0.018$ & 0 / 2 & 0.279 & $-0.012$ & 1 / 3 & $\sci{4.0}{-6}$\,$^{\dagger}$ \\
\midrule
Raw-feature baseline & $-0.019$ & n/a & n/a & $-0.019$ & n/a & n/a \\
\bottomrule
\end{tabular}
\end{table}
\section{Held-Out Commodities}
\label{app:heldout}

The macroeconomic-label comparison changes what defines a regime.
Held-out commodities provide a different extension by changing the
target asset class while retaining price-based regime labels. This
tests representations outside the asset classes used to train the
encoders, with selection separated from the final test period.

\subsection{Validation-selected commodities analysis}

For each method and seed, the clustering configuration in $G$ that
scores best on the 2021 to 2022 validation slice is held fixed, and that
configuration is scored on the 2023 to 2024 test slice, as shown in
Table~\ref{tab:heldout}. The raw-feature baseline is $k$-means with $k{=}3$ on the test
slice, scored at twenty seeds. Table~\ref{tab:heldout_vsraw} tests every
method against the baseline under this validation selection and
Table~\ref{tab:commodity_comparison} gives VICReg-TN's head-to-heads.

\begin{table}[h]
\centering
\small
\caption{Held-out commodities: mean test ARI $\pm$ standard deviation
over twenty seeds under the configuration selected on the validation
slice.}
\label{tab:heldout}
\begin{tabular}{lccc}
\toprule
Encoder & Crypto source & Equity source & FX source \\
\midrule
VICReg-TN & $0.138 \pm 0.053$ & $0.092 \pm 0.031$ & $0.130 \pm 0.074$ \\
SimCLR+VC & $0.096 \pm 0.045$ & $0.072 \pm 0.020$ & $0.125 \pm 0.056$ \\
Barlow Twins & $0.111 \pm 0.031$ & $0.159 \pm 0.038$ & $0.189 \pm 0.018$ \\
BYOL & $0.144 \pm 0.060$ & $0.133 \pm 0.052$ & $0.107 \pm 0.035$ \\
BT+VH & $0.112 \pm 0.023$ & $0.140 \pm 0.051$ & $0.197 \pm 0.018$ \\
VICReg-Hard & $-0.048 \pm 0.038$ & $0.098 \pm 0.028$ & $0.139 \pm 0.012$ \\
TS2Vec-style & $0.163 \pm 0.038$ & $0.134 \pm 0.065$ & $0.103 \pm 0.055$ \\
TNC-style & $0.177 \pm 0.064$ & $0.111 \pm 0.031$ & $0.113 \pm 0.036$ \\
VAE & $0.147 \pm 0.066$ & $0.225 \pm 0.056$ & $0.126 \pm 0.056$ \\
\bottomrule
\end{tabular}
\end{table}

\begin{table}[H]
\centering
\begin{minipage}[t]{0.49\textwidth}
\vspace{0pt}
\centering
\small
\begin{minipage}[t][8.5\baselineskip][t]{\linewidth}
\caption{Held-out commodities with validation-selected encoder
configurations and a $k{=}3$ $k$-means raw-feature baseline.
Mean ARI averages the three sources, and Wins counts seed-level direction
wins. Naive $p$ and $p_a$ are Fisher combinations before and after Brown
adjustment. $^{\dagger}$ marks Holm significance across nine methods.}
\label{tab:heldout_vsraw}
\end{minipage}
\begin{tabular}{lcccc}
\toprule
Encoder & Mean ARI & Wins & Naive $p$ & $p_a$ \\
\midrule
VICReg-TN & 0.120 & 0 / 3 & $1.000$ & $1.000$ \\
SimCLR+VC & 0.098 & 0 / 3 & $1.000$ & $1.000$ \\
Barlow Twins & 0.153 & 0 / 3 & $1.000$ & $1.000$ \\
BYOL & 0.128 & 0 / 3 & $1.000$ & $1.000$ \\
BT+VH & 0.150 & 0 / 3 & $0.997$ & $0.994$ \\
VICReg-Hard & 0.063 & 0 / 3 & $1.000$ & $1.000$ \\
TS2Vec-style & 0.133 & 0 / 3 & $1.000$ & $1.000$ \\
TNC-style & 0.134 & 0 / 3 & $1.000$ & $1.000$ \\
VAE & 0.166 & 1 / 3 & $0.318$ & $0.321$ \\
\midrule
Baseline & 0.200 & n/a & n/a & n/a \\
\bottomrule
\end{tabular}
\end{minipage}
\hfill
\begin{minipage}[t]{0.49\textwidth}
\vspace{0pt}
\centering
\small
\begin{minipage}[t][8.5\baselineskip][t]{\linewidth}
\caption{VICReg-TN comparisons on held-out commodities under
validation-selected clustering. One-sided Welch tests favouring
VICReg-TN are combined by Fisher's method across shared source assets.
$^{\dagger}$ marks Holm significance across eight pairs.}
\label{tab:commodity_comparison}
\end{minipage}
\begin{tabular}{lcc}
\toprule
VICReg-TN vs & Sources & $p$ \\
\midrule
SimCLR+VC & 3 & $0.002$\,$^{\dagger}$ \\
Barlow Twins & 3 & $0.304$ \\
BYOL & 3 & $0.502$ \\
BT+VH & 3 & $0.311$ \\
VICReg-Hard & 3 & $\sci{1.7}{-12}$\,$^{\dagger}$ \\
TS2Vec-style & 3 & $0.578$ \\
TNC-style & 3 & $0.748$ \\
VAE & 3 & $0.873$ \\
\bottomrule
\end{tabular}
\end{minipage}
\end{table}

\FloatBarrier
\Needspace{18\baselineskip}
\subsection{Sensitivity with the boundary-and-threshold rules}

To relate this extension to the revised evaluation procedure, we also
apply the boundary and threshold rules while retaining the previously
trained representations.

Table~\ref{tab:families_clean} extends the analysis with the
boundary-and-threshold rules
to commodities, REITs, and bonds. Each cell averages the three source
assets. Each entry lists mean ARI and the corresponding direction-win
count, separated by a slash. The label thresholds are fitted through 2020. We use a 29-window
boundary gap except on the commodities test slice, which immediately follows
the validation slice. Applying the boundary rule there as well does not
produce a method that beats the raw-feature baseline. This analysis reuses the mixed-horizon
checkpoints.

\begin{table}[H]
\centering
\footnotesize
\caption{Held-out families with the boundary-and-threshold rules.
Entries give mean ARI over three source assets followed by the
seed-level direction-win count against raw, separated by a slash. Scores use fixed $k{=}3$ $k$-means except
Val.-selected, which tests each seed's validation-chosen configuration.
$^{\dagger}$ marks Holm significance within each column's eleven-method
family, excluding the published-objective variants.}
\label{tab:families_clean}
\setlength{\tabcolsep}{4pt}
\begin{tabular}{@{}lccccc@{}}
\toprule
 & \multicolumn{3}{c}{Commodities} & & \\
\cmidrule(lr){2-4}
Method & Validation & Test & Val.-selected & REIT & Bond \\
\midrule
VICReg-TN & 0.172\,$^{\dagger}$ / 3 & 0.142 / 0 & 0.128 / 0 & 0.460\,$^{\dagger}$ / 3 & 0.043 / 1 \\
SimCLR+VC & 0.254\,$^{\dagger}$ / 3 & 0.119 / 0 & 0.117 / 0 & 0.449\,$^{\dagger}$ / 3 & 0.087\,$^{\dagger}$ / 1 \\
Barlow Twins & 0.216\,$^{\dagger}$ / 3 & 0.114 / 0 & 0.124 / 0 & 0.335\,$^{\dagger}$ / 1 & 0.172\,$^{\dagger}$ / 2 \\
BYOL & 0.277\,$^{\dagger}$ / 3 & 0.109 / 0 & 0.118 / 0 & 0.530\,$^{\dagger}$ / 3 & 0.210\,$^{\dagger}$ / 3 \\
BT+VH & 0.220\,$^{\dagger}$ / 3 & 0.120 / 0 & 0.127 / 0 & 0.327\,$^{\dagger}$ / 1 & 0.179\,$^{\dagger}$ / 2 \\
VICReg-Hard & 0.109\,$^{\dagger}$ / 2 & 0.027 / 0 & 0.044 / 0 & 0.433\,$^{\dagger}$ / 2 & 0.098\,$^{\dagger}$ / 2 \\
TS2Vec-style & 0.177\,$^{\dagger}$ / 3 & 0.111 / 0 & 0.106 / 0 & 0.298 / 0 & 0.051 / 0 \\
TNC-style & 0.247\,$^{\dagger}$ / 3 & 0.124 / 0 & 0.116 / 0 & 0.490\,$^{\dagger}$ / 2 & 0.097\,$^{\dagger}$ / 1 \\
VAE & 0.118\,$^{\dagger}$ / 3 & 0.119 / 0 & 0.124 / 0 & 0.371\,$^{\dagger}$ / 1 & 0.062 / 0 \\
TS-TCC-style & 0.063\,$^{\dagger}$ / 3 & 0.051 / 0 & 0.068 / 0 & 0.170 / 0 & 0.039 / 0 \\
CoST-style & 0.057\,$^{\dagger}$ / 3 & 0.090 / 0 & 0.108 / 0 & 0.197 / 0 & 0.070 / 1 \\
TS-TCC / published obj. & 0.232 / 3 & 0.079 / 0 & 0.083 / 0 & 0.334 / 1 & 0.109 / 3 \\
CoST published obj. & 0.157 / 3 & 0.062 / 0 & 0.067 / 0 & 0.207 / 0 & 0.091 / 2 \\
\midrule
Baseline & $-0.007$ & 0.197 & 0.197 & 0.357 & 0.057
 \\
\bottomrule
\end{tabular}
\end{table}

\section{Additional Label Families in the Earlier Evaluation}
\label{app:families}

The preceding analysis reports the revised protocol on the additional
asset classes. This complementary sweep examines how the label-family
comparison moves across selectors in the earlier evaluation. It uses
the mixed-horizon evaluation, without the
boundary rule and with thresholds fitted on the full available period.
Table~\ref{tab:famsel} compares VICReg-TN with raw across label families
and selectors at twenty seeds. For the label-free quality criteria,
each seed's configuration is chosen separately. ``Aggregate'' combines
the crypto, equity and FX target families. ``Commodities'' uses the
held-out test cells described in Appendix~\ref{app:heldout}.

VICReg-TN improves on raw under most selectors on the REIT family,
whereas no selector establishes an improvement on bonds.
Table~\ref{tab:reit_vsraw} extends the fixed-selector REIT comparison
to every encoder. The aggregate GMM comparison reflects reduced raw
performance, as discussed in Section~\ref{sec:flip}, while significant
Sahm comparisons occur near zero ARI, as detailed in Appendix~\ref{app:sahm}. These tests condition
on the stored evaluation periods.

\begin{table}[h]
\centering
\small
\caption{Selector sensitivity in the earlier mixed-horizon evaluation.
Entries are Brown-adjusted Fisher $p$ values for VICReg-TN against raw
features, with the same selector applied to both. Aggregate combines
the crypto, equity and FX target families.}
\label{tab:famsel}
\begin{tabular}{lccccc}
\toprule
Selector & Aggregate & Sahm & Commodities & REIT & Bond \\
\midrule
$k{=}3$ $k$-means, fixed & $0.267$ & $\sci{1.4}{-12}$ & $1.000$ & $\sci{4.9}{-6}$ & $1.000$ \\
$k{=}3$ GMM & $<10^{-15}$ & $\sci{1.5}{-11}$ & $0.710$ & $\sci{2.4}{-7}$ & $1.000$ \\
Silhouette & $0.934$ & $<10^{-15}$ & $0.759$ & $\sci{9.6}{-5}$ & $1.000$ \\
Calinski-Harabasz & $0.965$ & $\sci{1.4}{-12}$ & $1.000$ & $\sci{7.8}{-6}$ & $1.000$ \\
Davies-Bouldin & $1.000$ & $<10^{-15}$ & $0.994$ & $0.118$ & $1.000$ \\
Label-aware & $\sci{2.6}{-6}$ & $\sci{1.8}{-9}$ & $0.996$ & $0.758$ & $1.000$ \\
\bottomrule
\end{tabular}
\end{table}

\begin{table}[h]
\centering
\small
\caption{REIT comparison against raw features in the earlier mixed-horizon evaluation under
fixed $k{=}3$ $k$-means. Mean ARI averages available source assets.
Wins counts seed-level direction wins over evaluated sources, with FX
omitted in the two-source rows. Naive $p$ and $p_a$ are unadjusted and
Brown-adjusted Fisher combinations. $^{\dagger}$ marks Holm significance
among nine encoders.}
\label{tab:reit_vsraw}
\begin{tabular}{lcccc}
\toprule
Encoder & Mean ARI & Wins & Naive $p$ & $p_a$ \\
\midrule
VICReg-TN & 0.244 & 2 / 3 & $\sci{8.2}{-7}$ & $\sci{4.9}{-6}$\,$^{\dagger}$ \\
SimCLR+VC & 0.232 & 1 / 2 & $0.014$ & $0.020$ \\
Barlow Twins & 0.248 & 2 / 2 & $\sci{8.8}{-12}$ & $\sci{8.8}{-12}$\,$^{\dagger}$ \\
BYOL & 0.328 & 3 / 3 & $<10^{-15}$ & $\sci{2.5}{-12}$\,$^{\dagger}$ \\
BT+VH & 0.335 & 3 / 3 & $<10^{-15}$ & $<10^{-15}$\,$^{\dagger}$ \\
VICReg-Hard & 0.386 & 3 / 3 & $<10^{-15}$ & $<10^{-15}$\,$^{\dagger}$ \\
TS2Vec-style & 0.192 & 1 / 3 & $0.014$ & $0.019$ \\
TNC-style & 0.236 & 1 / 3 & $\sci{8.8}{-6}$ & $\sci{2.3}{-5}$\,$^{\dagger}$ \\
VAE & 0.150 & 0 / 3 & $0.886$ & $0.866$ \\
\midrule
Baseline & 0.218 & n/a & n/a & n/a \\
\bottomrule
\end{tabular}
\end{table}

\section{Linear Probe}
\label{app:probe}

A clustering score depends both on the information in a representation
and on how that information is partitioned. A linear probe changes
the readout to ask whether regime labels are accessible through a
supervised linear decision rule on the frozen representation.

\subsection{Probe in the earlier mixed-horizon evaluation}

\begin{figure}[H]
\centering
\includegraphics[width=0.88\linewidth]{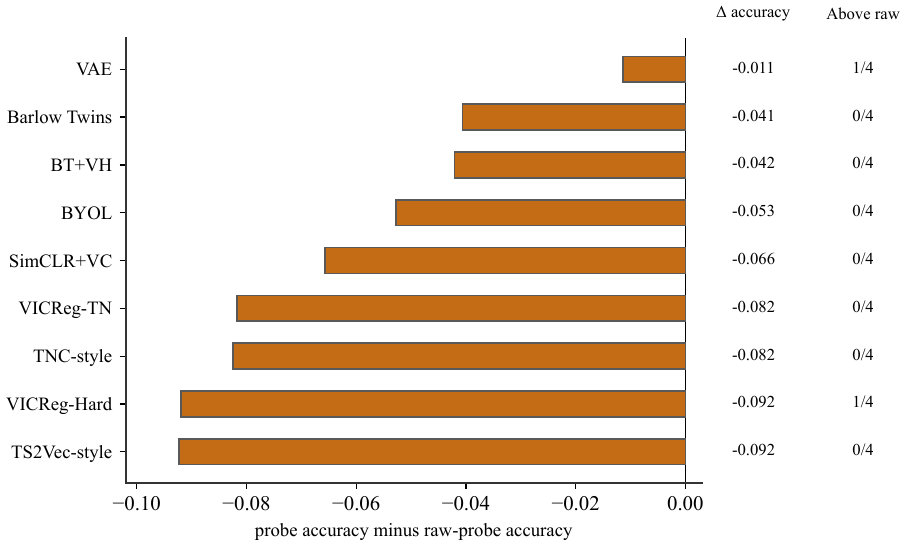}
\caption{Linear-probe accuracy relative to raw features in the earlier
mixed-horizon evaluation, averaged across transfer directions with equity
or FX targets. Above raw counts directions with higher encoder accuracy.}
\label{fig:probe}
\end{figure}

The probe is a multinomial logistic regression from the fixed embedding
to the target labels. It is trained on the 2021 to 2022 slice of the
target's evaluation window and tested on the 2023 to 2024 slice, so it
uses only the equity and FX targets, whose evaluation windows span both
slices, in the four directions that end there. The crypto evaluation
window starts in 2023 and has no training slice. The VIX-level labels of
the 2021 to 2022 slice hold two of the three classes. The raw-feature probe is
the same regression on the thirty raw features, which has no seed
variance. Against this baseline, an encoder's mean accuracy is compared per
direction by its sign. In head-to-heads the per-seed accuracies of two
encoders are compared by one-sided Welch tests, combined by Fisher over
the four directions using the naive combination without Brown adjustment.
Figure~\ref{fig:probe} shows the mean accuracy differences. Every method
is less accurate than the raw-feature probe on average, although two
methods exceed it in one direction. The head-to-head results also depend
on the direction-level combination rather than the mean.

\subsection{Probe with the boundary-and-threshold rules}

Table~\ref{tab:probe_clean} repeats the probe after applying the boundary
rule to the training slice and fitting the FX labels
through 2020, for all thirteen encoders. It retains the earlier
mixed-horizon checkpoints.

\begin{table}[h]
\centering
\small
\caption{Linear probe with the boundary-and-threshold rules.
Accuracy and macro-F1 average the four equity- and FX-target directions.
Above baseline counts positive mean-accuracy differences from raw, and
Significant counts one-sided Welch rejections. Fisher $p$ combines these
tests without dependence adjustment.}
\label{tab:probe_clean}
\begin{tabular}{lccccc}
\toprule
Encoder & Accuracy & Macro-F1 & Above baseline & Significant & Fisher $p$ \\
\midrule
VICReg-TN & 0.518 & 0.423 & 1 / 4 & 0 / 4 & $0.964$ \\
SimCLR+VC & 0.547 & 0.390 & 0 / 4 & 0 / 4 & $1.000$ \\
Barlow Twins & 0.580 & 0.434 & 1 / 4 & 1 / 4 & $\sci{1.2}{-4}$ \\
BYOL & 0.562 & 0.424 & 2 / 4 & 1 / 4 & $0.044$ \\
BT+VH & 0.579 & 0.432 & 1 / 4 & 1 / 4 & $\sci{6.6}{-4}$ \\
VICReg-Hard & 0.554 & 0.425 & 1 / 4 & 0 / 4 & $0.743$ \\
TS2Vec-style & 0.508 & 0.392 & 0 / 4 & 0 / 4 & $1.000$ \\
TNC-style & 0.530 & 0.396 & 0 / 4 & 0 / 4 & $1.000$ \\
VAE & 0.604 & 0.463 & 1 / 4 & 1 / 4 & $\sci{8.5}{-10}$ \\
TS-TCC-style & 0.545 & 0.393 & 0 / 4 & 0 / 4 & $1.000$ \\
CoST-style & 0.522 & 0.382 & 0 / 4 & 0 / 4 & $1.000$ \\
TS-TCC, published obj. & 0.526 & 0.398 & 0 / 4 & 0 / 4 & $1.000$ \\
CoST published obj. & 0.471 & 0.350 & 0 / 4 & 0 / 4 & $1.000$ \\
\midrule
Baseline & 0.592 & 0.433 & n/a & n/a & n/a
 \\
\bottomrule
\end{tabular}
\end{table}

\section{Synthetic Known-Regime Study}
\label{app:synthetic}
The real-market studies compare representations against constructed
regime proxies. Known generating states allow a more direct examination
of which types of change the window features and learned representations
can capture. The synthetic design therefore separates changes in
volatility and correlation levels from changes in temporal dynamics.

The study of Section~\ref{sec:synthetic} generates thirty-asset daily
return series from a one-factor model $r_{it} = \beta_i f_t + e_{it}$
with loadings $\beta_i \sim \mathrm{Uniform}(0.6, 1.4)$ drawn once, over
a three-state Markov chain with mean sojourn 60 trading days and uniform
exit probabilities, for 3{,}000 retained days after a 100-day burn-in,
using data seed 7. The regimes are calm, stress and recovery. In the
\emph{level} family each regime has its own average daily volatility and
average pairwise correlation, obtained by solving the factor and
idiosyncratic variances for the targets. At separation 1, calm has
volatility 0.8\% and correlation 0.25, stress has 2.0\% and 0.60, and
recovery has 1.2\% and 0.40. The
separation $\delta \in \{0.5, 1, 2\}$ scales the calm and stress gaps
around the recovery anchor. In the \emph{dynamics} family all three
regimes share the recovery level and differ only in the factor's first-order autoregressive
coefficient, $0$ in calm, $+0.5\delta$ in stress and $-0.5\delta$ in
recovery, capped at $\pm 0.95$, with the innovation variance rescaled so
that the factor's stationary variance, and hence the marginal volatility
and correlation, is the same in every regime. The thirty summary features
defined in Section~\ref{sec:prelim} are level statistics of a thirty-day
window, so the level family is a signal the raw-feature baseline sees
directly, while the dynamics family has
no difference in marginal level across regimes.

Each series is passed through the same feature extractor, split into the
first 2{,}000 days for training and the last 1{,}000 for evaluation, and
labelled by the generating chain. This yields 1{,}970 training windows and
999 evaluation windows, with all three regimes present in evaluation. All thirteen methods are trained on
the training windows at five seeds, 42 to 46, with the benchmark defaults
of Appendix~\ref{app:methods}. Training uses 100 epochs, embedding width
16 and learning rate $10^{-3}$. The six joint-embedding MLP encoders share
the same backbone, temporal positives, augmentation and training procedure and differ only
in the loss. All are
evaluated in-domain after applying the boundary rule, with the fixed
selector and raw-feature $k$-means at five seeds. The five training
seeds share the same generated series and therefore are not independent
replications of the data-generating process. Table~\ref{tab:synthetic}
reports descriptive seed means without significance marks.

\begin{table}[t]
\caption{Synthetic regime identification at the separation values $\delta$
shown in the column headings. Encoder entries are descriptive seed-mean ARI
differences from raw under fixed selection, using one generated series
per condition. The last two rows give absolute raw ARI. Level regimes
differ in volatility and correlation, while dynamics regimes differ only
in factor autocorrelation.}
\label{tab:synthetic}
\centering
\small
\setlength{\tabcolsep}{4pt}
\begin{tabular}{@{}lcccccc@{}}
\toprule
 & \multicolumn{3}{c}{Level} & \multicolumn{3}{c}{Dynamics} \\
\cmidrule(lr){2-4}\cmidrule(lr){5-7}
Encoder & 0.5 & 1 & 2 & 0.5 & 1 & 2 \\
\midrule
VICReg-TN & $-0.048$ & $-0.082$ & $-0.243$ & $+0.034$ & $+0.042$ & $+0.069$ \\
SimCLR+VC & $-0.018$ & $-0.085$ & $-0.133$ & $+0.029$ & $+0.036$ & $+0.072$ \\
Barlow Twins & $+0.041$ & $+0.071$ & $+0.051$ & $+0.011$ & $+0.033$ & $+0.074$ \\
BYOL & $+0.026$ & $-0.054$ & $-0.080$ & $+0.003$ & $+0.001$ & $+0.054$ \\
BT+VH & $+0.045$ & $+0.069$ & $+0.051$ & $+0.007$ & $+0.034$ & $+0.076$ \\
VICReg-Hard & $+0.042$ & $+0.036$ & $+0.010$ & $+0.007$ & $+0.025$ & $+0.111$ \\
TS2Vec-style & $+0.023$ & $-0.093$ & $-0.068$ & $+0.022$ & $+0.056$ & $-0.017$ \\
TNC-style & $+0.072$ & $+0.134$ & $+0.092$ & $+0.009$ & $+0.098$ & $+0.130$ \\
VAE & $-0.012$ & $-0.041$ & $-0.033$ & $+0.003$ & $+0.009$ & $+0.023$ \\
TS-TCC-style & $-0.171$ & $-0.274$ & $-0.345$ & $+0.004$ & $+0.008$ & $+0.009$ \\
CoST-style & $-0.051$ & $-0.198$ & $-0.214$ & $+0.011$ & $+0.007$ & $+0.008$ \\
TS-TCC, published obj. & $-0.063$ & $-0.124$ & $-0.072$ & $+0.019$ & $+0.063$ & $+0.019$ \\
CoST published obj. & $-0.064$ & $-0.150$ & $-0.223$ & $+0.003$ & $+0.006$ & $-0.002$ \\
\midrule
Baseline ARI & 0.174 & 0.279 & 0.359 & $-0.001$ & 0.001 & 0.050 \\
Baseline ARI, label-aware & 0.178 & 0.320 & 0.495 & 0.011 & 0.006 & 0.214
 \\
\bottomrule
\end{tabular}
\end{table}

\begin{figure}[H]
\centering
\includegraphics[width=0.96\textwidth]{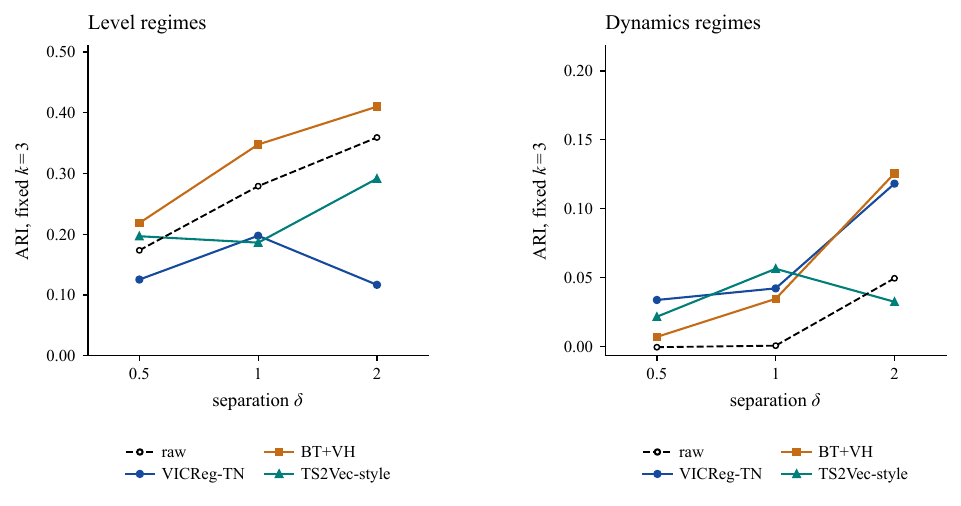}
\caption{Synthetic regime recovery as separation $\delta$ increases.
The left panel varies volatility and correlation levels, while the right
panel varies factor autocorrelation alone. Curves show descriptive seed-mean ARI under
$k$-means with $k{=}3$.}
\label{fig:synthetic}
\end{figure}

Table~\ref{tab:synthetic} and Figure~\ref{fig:synthetic} show how the
comparison changes with the type of regime signal. For level regimes,
the raw-feature baseline improves as separation grows.
Some methods exceed it at every separation, while others trail it
throughout. Label-aware selection reverses the comparison for VICReg-TN,
showing that the conclusion also depends on the clustering rule. For
dynamics regimes, most method means are higher than the baseline, but the
absolute scores remain small. Independent generated series
would be required before interpreting either pattern as population-level
evidence.

\FloatBarrier

\section{Effective Rank and Checkpoint Selection}
\label{app:er}

The preceding studies evaluate completed representations. A separate
question is whether training can be stopped using an unlabelled
property of the embedding. The tuning experiment in
Section~\ref{sec:tuning} uses effective rank to exclude collapsed
configurations. It is the exponential of the entropy of the normalised
eigenvalues of the embedding covariance on the recorded embeddings.
The training trajectories below instead use the singular values of
the centred embedding matrix.

We examine effective rank as an early-stopping signal by training
five encoders on an extended schedule with periodic checkpoints.
These are additional single-seed runs at seed 42, scored without the boundary rule,
and they are not the benchmark's configurations: SimCLR+VC at temperature 0.5,
BT+VH, VICReg-Hard and BYOL ran 500 to 1{,}400 epochs on the thirty features
with neighbourhood 5, and VICReg-TN ran 1{,}000 epochs on the 63-dimensional
feature vector with loss weights 10, 25 and 5 and neighbourhood 30.
The results do not support using effective rank as an early-stopping
signal: all five methods have an ARI drop of at least 0.02 without a
relative effective-rank drop above 0.05, as shown in Table~\ref{tab:er}. A stopping
rule that selects the first checkpoint whose effective rank falls below
0.99 of its running peak provides no consistent improvement over the
last epoch. This stopping comparison uses mean ARI across the six
clustering configurations. These runs support using effective rank as a collapse
filter, with limited evidence for checkpoint selection.

\begin{table}[h]
\centering
\small
\caption{Extended-training trajectories for effective rank, denoted ER, and ARI.
ER drop is the relative peak-to-final decrease, and $\Delta$ ARI is the
absolute peak-to-final decrease. Rank corr.\ compares the two trajectories.
ARI is maximised over the six clustering configurations at each checkpoint.}
\label{tab:er}
\begin{tabular}{lcccccc}
\toprule
Encoder & Peak-ARI epoch & Peak-ER epoch & ER drop & $\Delta$ ARI & Rank corr. \\
\midrule
VICReg-TN & 100 & 200 & 4.3\% & 0.128 & $+0.491$ \\
SimCLR+VC & 200 & 200 & 1.0\% & 0.022 & $+0.700$ \\
BT+VH & 400 & 1{,}200 & 0.1\% & 0.054 & $-0.357$ \\
VICReg-Hard & 200 & 500 & 0.0\% & 0.028 & $+0.100$ \\
BYOL & 800 & 1{,}400 & 0.0\% & 0.058 & $+0.357$ \\
\bottomrule
\end{tabular}
\end{table}

\section{Notation and Additional Tables}
\label{app:extra_tables}
The following tables provide notation and complete numerical comparisons
for the main and appendix plots. They collect training and clustering
selection, temporal-dependence diagnostics, calibration, objective variants,
and protocol and period comparisons.

Table~\ref{tab:notation} collects the notation of Section~\ref{sec:prelim}.
Table~\ref{tab:hpo} holds the tuning results of Section~\ref{sec:tuning},
Table~\ref{tab:ksweep} the selector sensitivity on the aggregate and
Table~\ref{tab:ess} the temporal-dependence diagnostic introduced in
Section~\ref{sec:inference}.
The clustering sweep uses the stored grid for $k\in\{3,4,5\}$ and
re-extracts all 54 method--direction cells for $k\in\{2,6\}$.
The dependence diagnostic uses clustering seed 42.

\begin{table}[h]
\caption{Notation.}
\label{tab:notation}
\centering
\footnotesize
\setlength{\tabcolsep}{3pt}
\begin{tabular}{@{}ll@{}}
\toprule
Symbol & Meaning \\
\midrule
$x_u \in \mathbb{R}^{30}$ & feature vector of the window ending on day $u$ \\
$f_\theta,\ z_u$ & encoder and its embedding $z_u = f_\theta(x_u) \in \mathbb{R}^{16}$ \\
$s \to t$ & transfer direction from source $s$ to target $t$, with $K{=}6$ \\
$\ell,\ y^{\ell}_u$ & label family and its label for window $u$ \\
$G$ & grid $\{\text{k-means}, \text{GMM}\} \times \{3,4,5\}$ \\
$(m,k)$ & clustering method and number of clusters, one element of $G$ \\
$W{=}30$ & window length in days \\
$Z,\ \hat c^{\,g}$ & target embeddings of one seed and clusters under configuration $g$ \\
$\pi$ & selector, a map from the grid to one configuration \\
$\mathrm{ARI}$ & adjusted Rand index~\citep{hubert1985ari} \\
$n{=}20$ & seeds per method, direction and selector on real data \\
$p_i,\ \chi^2,\ p_a$ & per-direction $p$, Fisher statistic, Brown-adjusted $p$ \\
$\rho_{ij},\ c,\ \mathrm{df}_{\mathrm{eff}}$ & seed correlation of two directions, Brown scale and d.f. \\
$R,\ L$ & block-bootstrap replicates and block length \\
$M,\ B$ & tuning budget per source asset and batch size \\
$N,\ N_{\mathrm{eff}},\ \rho_1$ & nominal and effective windows, lag-1 autocorr. \\
\bottomrule
\end{tabular}
\end{table}

\begin{table}[H]
\centering
\begin{minipage}[t]{0.49\textwidth}
\vspace{0pt}
\begin{minipage}[t][10.5\baselineskip][t]{\linewidth}
\caption{Budget-matched tuning in the earlier mixed-horizon evaluation
under fixed selection. Def., LF and SL
count direction wins out of six for default, label-free and source-label
tuning. SL uses LF where the source-label criterion is undefined.
$p_a$ compares LF with raw using Brown-adjusted Fisher combination.
$\Delta$ is LF-minus-default mean ARI. Oracle is the source-mean test-ARI
gain of the best configuration over default.}
\label{tab:hpo}
\end{minipage}
\centering
\scriptsize
\setlength{\tabcolsep}{3.5pt}
\begin{tabular}{@{}lcccccc@{}}
\toprule
Encoder & Def. & LF & $p_a$ & $\Delta$ & SL & Oracle \\
\midrule
VICReg-TN & 1 & 1 & $0.276$ & $-0.003$ & 0 & $+0.008$ \\
SimCLR+VC & 1 & 1 & $0.008$ & $-0.000$ & 1 & $+0.020$ \\
Barlow Twins & 1 & 1 & $0.324$ & $-0.024$ & 0 & $+0.013$ \\
BYOL & 0 & 0 & $1.000$ & $-0.010$ & 0 & $+0.060$ \\
BT+VH & 1 & 0 & $0.997$ & $-0.017$ & 0 & $+0.034$ \\
VICReg-Hard & 0 & 0 & $1.000$ & $-0.026$ & 0 & $+0.073$ \\
TS2Vec-style & 0 & 0 & $0.963$ & $-0.008$ & 1 & $+0.052$ \\
TNC-style & 0 & 0 & $0.911$ & $+0.026$ & 0 & $+0.037$ \\
VAE & 0 & 2 & $0.004$ & $+0.011$ & 1 & $+0.050$ \\
TS-TCC-style & 0 & 0 & $1.000$ & $+0.010$ & 0 & $+0.050$ \\
CoST-style & 0 & 0 & $1.000$ & $-0.006$ & 0 & $+0.040$ \\
\bottomrule
\end{tabular}
\end{minipage}
\hfill
\begin{minipage}[t]{0.49\textwidth}
\vspace{0pt}
\begin{minipage}[t][10.5\baselineskip][t]{\linewidth}
\caption{Clustering sensitivity in the earlier mixed-horizon aggregate.
Max is the largest significant-direction count among nine encoders,
out of six directions. A uniquely best encoder is named beside its count.
VICReg-TN $p_a$ is its Brown-adjusted Fisher comparison with raw
under the same clustering rule.}
\label{tab:ksweep}
\end{minipage}
\centering
\footnotesize
\setlength{\tabcolsep}{3pt}
\begin{tabular}{@{}ccccc@{}}
\toprule
& \multicolumn{2}{c}{$k$-means} & \multicolumn{2}{c}{GMM} \\
\cmidrule(lr){2-3}\cmidrule(lr){4-5}
$k$ & Max & VICReg-TN $p_a$ & Max & VICReg-TN $p_a$ \\
\midrule
2 & 4/6, BT+VH & $<10^{-15}$ & 4/6 & $<10^{-15}$ \\
\textbf{3} & \textbf{1/6} & $\mathbf{0.267}$ & 4/6 & $<10^{-15}$ \\
4 & 2/6 & $\sci{2.1}{-4}$ & 4/6 & $<10^{-15}$ \\
5 & 3/6 & $0.048$ & 5/6 & $<10^{-15}$ \\
6 & 3/6 & $0.249$ & 5/6 & $<10^{-15}$ \\
\bottomrule
\end{tabular}
\end{minipage}
\end{table}

\begin{table}[H]
\caption{Temporal-dependence diagnostics in the earlier mixed-horizon
evaluation. Train/evaluation pairs give nominal windows $N$,
first-order autoregressive effective windows $N_{\mathrm{eff}}$ and regime episodes. Baseline ARI
uses fixed $k{=}3$ clustering on overlapping windows or disjoint phase
samples. Phase columns summarise the 30 starting offsets, each sampled
at stride 30.}
\label{tab:ess}
\centering
\footnotesize
\setlength{\tabcolsep}{5pt}
\begin{tabular}{@{}llcccccc@{}}
\toprule
& & \multicolumn{3}{c}{Train / evaluation} & \multicolumn{3}{c}{Baseline ARI, evaluation split} \\
\cmidrule(lr){3-5}\cmidrule(lr){6-8}
Asset & Family & $N$ & $N_{\mathrm{eff}}$ & Episodes & Overlapping & Phase mean $\pm$ sd & Phase min / max \\
\midrule
Crypto & manual crypto & 1431 / 730 & 80 / 43 & 3 / 2 & 0.097 & $0.028 \pm 0.101$ & $-0.075$ / $0.273$ \\
Equity & VIX-level & 1481 / 1003 & 28 / 55 & 147 / 92 & 0.132 & $0.115 \pm 0.064$ & $-0.007$ / $0.231$ \\
FX & FX volatility & 1536 / 1041 & 71 / 56 & 99 / 58 & 0.250 & $0.240 \pm 0.093$ & $0.089$ / $0.485$ \\
\bottomrule
\end{tabular}
\end{table}

\begin{table}[H]
\caption{Seed-level calibration for nine encoders in the aligned 2023--24 evaluation.
Values are percentage rejections at $\alpha{=}0.05$ under centred joint
swaps, a fitted parametric null and a null with a shared seed shock.
Adjusted denotes Brown-adjusted Fisher tests. Power is the adjusted
test's detection rate for the indicated ARI increase in equity $\to$ FX.}
\label{tab:calibration}
\centering
\small
\setlength{\tabcolsep}{5pt}
\begin{tabular}{@{}lccccccc@{}}
\toprule
 & \multicolumn{2}{c}{Joint swap} & \multicolumn{2}{c}{Parametric} & Coupled & \multicolumn{2}{c}{Power} \\
\cmidrule(lr){2-3}\cmidrule(lr){4-5}\cmidrule(lr){7-8}
Encoder & naive & adjusted & naive & adjusted & adjusted & 0.02 & 0.05 \\
\midrule
VICReg-TN & 7.2 & 5.8 & 6.4 & 4.5 & 7.6 & 26.1 & 93.7 \\
SimCLR+VC & 4.5 & 2.6 & 5.5 & 4.0 & 7.2 & 83.7 & 100.0 \\
Barlow Twins & 3.4 & 2.8 & 7.0 & 4.9 & 7.9 & 89.4 & 100.0 \\
BYOL & 3.1 & 2.4 & 4.6 & 4.0 & 7.1 & 25.6 & 95.0 \\
BT+VH & 10.9 & 7.0 & 7.4 & 5.0 & 6.6 & 93.8 & 100.0 \\
VICReg-Hard & 2.2 & 1.4 & 7.8 & 5.1 & 6.3 & 100.0 & 100.0 \\
TS2Vec-style & 8.8 & 5.5 & 6.8 & 4.8 & 7.2 & 55.2 & 99.8 \\
TNC-style & 2.5 & 2.3 & 5.3 & 4.2 & 6.4 & 34.7 & 98.5 \\
VAE & 8.5 & 5.8 & 6.2 & 5.1 & 8.0 & 29.0 & 96.0 \\
\midrule
Mean & 5.7 & 3.9 & 6.3 & 4.6 & 7.2 &  &
 \\
\bottomrule
\end{tabular}
\end{table}

\begin{table}[h]
\centering
\small
\caption{Adapted and published-objective variants under fixed selection
and the boundary-and-threshold rules. Mean ARI averages six transfer
directions in the first three rows and two equity--FX directions in the
final two. Direction wins and Brown-adjusted Fisher $p_a$
compare each published variant with raw features. $^{\dagger}$ marks
Holm significance within the thirteen-method sensitivity family.}
\label{tab:faithful}
\setlength{\tabcolsep}{4pt}
\begin{tabular}{@{}lccccccc@{}}
\toprule
 & & \multicolumn{3}{c}{TS-TCC} & \multicolumn{3}{c}{CoST} \\
\cmidrule(lr){3-5}\cmidrule(lr){6-8}
Setting & Baseline & Adapted & Published obj. & Wins / $p_a$ & Adapted & Published obj. & Wins / $p_a$ \\
\midrule
Mixed horizon & 0.187 & 0.085 & 0.101 & 0 / $0.951$ & 0.114 & 0.100 & 0 / $0.970$ \\
2021--22 & 0.314 & 0.190 & 0.274 & 1 / $0.031$ & 0.246 & 0.232 & 1 / $0.417$ \\
2023--24 & 0.070 & 0.023 & 0.071 & 3 / $\sci{2.1}{-13}$\,$^{\dagger}$ & 0.033 & 0.040 & 0 / $0.778$ \\
2019--20 & 0.224 & 0.193 & 0.237 & 1 / $<10^{-15}$\,$^{\dagger}$ & 0.223 & 0.209 & 1 / $\sci{2.4}{-7}$\,$^{\dagger}$ \\
2017--18 & 0.394 & 0.184 & 0.159 & 0 / $1.000$ & 0.344 & 0.157 & 0 / $1.000$
 \\
\bottomrule
\end{tabular}
\end{table}

\begin{table}[H]
\caption{Boundary and label-threshold ablation under the fixed selector.
Above baseline counts methods with positive six-direction mean effects.
Holm counts corrected combined-test rejections across eleven methods,
and Max wins is the largest seed-level direction-win count. The final
column gives VICReg-TN's mean ARI and direction wins.}
\label{tab:protocol_ablation}
\centering
\footnotesize
\setlength{\tabcolsep}{2.5pt}
\begin{tabular}{@{}llcccccccc@{}}
\toprule
Setting & Variant & \multicolumn{3}{c}{Baseline ARI} & Above baseline & Holm & Max wins & VICReg-TN \\
\cmidrule(lr){3-5}
 & & crypto & equity & FX & out of 11 & out of 11 & out of 6 & mean / wins \\
\midrule
Mixed horizon & Full-period, no exclusion & 0.093 & 0.132 & 0.272 & 0 & 1 & 1 & 0.129 / 1 \\
Mixed horizon & Training-period, no exclusion & 0.093 & 0.132 & 0.364 & 0 & 2 & 1 & 0.137 / 1 \\
Mixed horizon & Full-period, exclusion & 0.033 & 0.144 & 0.293 & 0 & 6 & 2 & 0.127 / 2 \\
Mixed horizon & Training-period, exclusion & 0.033 & 0.144 & 0.385 & 0 & 6 & 2 & 0.134 / 2 \\
\midrule
2023--24 & Full-period, no exclusion & 0.093 & 0.036 & 0.195 & 0 & 8 & 3 & 0.093 / 3 \\
2023--24 & Training-period, no exclusion & 0.093 & 0.036 & 0.169 & 0 & 8 & 3 & 0.089 / 3 \\
2023--24 & Full-period, exclusion & 0.033 & 0.031 & 0.160 & 3 & 6 & 4 & 0.096 / 4 \\
2023--24 & Training-period, exclusion & 0.033 & 0.031 & 0.147 & 3 & 6 & 4 & 0.092 / 4
 \\
\bottomrule
\end{tabular}
\end{table}

\begin{table}[H]
\centering
\begin{minipage}[t]{0.40\textwidth}
\vspace{0pt}
\begin{minipage}[t][5.5\baselineskip][t]{\linewidth}
\caption{All-asset transfer effects. Each block averages seed-mean ARI
differences from raw over six directions. Mean averages the two blocks.}
\label{tab:period}
\end{minipage}
\centering
\small
\setlength{\tabcolsep}{5pt}
\begin{tabular}{@{}lccc@{}}
\toprule
Method & 2021--22 & 2023--24 & Mean \\
\midrule
VICReg-TN & $-0.094$ & $+0.022$ & $-0.036$ \\
SimCLR+VC & $-0.101$ & $+0.018$ & $-0.041$ \\
Barlow Twins & $-0.086$ & $-0.024$ & $-0.055$ \\
BYOL & $-0.043$ & $-0.067$ & $-0.055$ \\
BT+VH & $-0.080$ & $-0.025$ & $-0.053$ \\
VICReg-Hard & $-0.016$ & $-0.066$ & $-0.041$ \\
TS2Vec-style & $-0.060$ & $+0.012$ & $-0.024$ \\
TNC-style & $-0.031$ & $-0.007$ & $-0.019$ \\
VAE & $-0.037$ & $-0.019$ & $-0.028$ \\
TS-TCC-style & $-0.125$ & $-0.047$ & $-0.086$ \\
CoST-style & $-0.068$ & $-0.037$ & $-0.053$
 \\
\bottomrule
\end{tabular}
\end{minipage}
\hfill
\begin{minipage}[t]{0.58\textwidth}
\vspace{0pt}
\begin{minipage}[t][5.5\baselineskip][t]{\linewidth}
\caption{Equity--FX transfer effects. Each block averages seed-mean ARI
differences from raw over two directions. Mean averages the four blocks.}
\label{tab:period_equityfx}
\end{minipage}
\centering
\small
\setlength{\tabcolsep}{5pt}
\begin{tabular}{@{}lccccc@{}}
\toprule
Method & 2017--18 & 2019--20 & 2021--22 & 2023--24 & Mean \\
\midrule
VICReg-TN & $-0.132$ & $+0.011$ & $-0.101$ & $+0.009$ & $-0.053$ \\
SimCLR+VC & $-0.188$ & $+0.020$ & $-0.105$ & $+0.024$ & $-0.062$ \\
Barlow Twins & $-0.168$ & $+0.154$ & $+0.005$ & $+0.012$ & $+0.000$ \\
BYOL & $-0.045$ & $+0.150$ & $-0.064$ & $-0.031$ & $+0.003$ \\
BT+VH & $-0.190$ & $+0.161$ & $+0.010$ & $+0.013$ & $-0.002$ \\
VICReg-Hard & $-0.290$ & $-0.012$ & $-0.040$ & $-0.035$ & $-0.094$ \\
TS2Vec-style & $-0.081$ & $-0.007$ & $-0.126$ & $-0.026$ & $-0.060$ \\
TNC-style & $-0.089$ & $-0.008$ & $-0.046$ & $-0.022$ & $-0.041$ \\
VAE & $+0.006$ & $+0.003$ & $-0.001$ & $+0.006$ & $+0.003$ \\
TS-TCC-style & $-0.209$ & $-0.031$ & $-0.235$ & $-0.071$ & $-0.137$ \\
CoST-style & $-0.050$ & $-0.001$ & $-0.151$ & $-0.063$ & $-0.066$
 \\
\bottomrule
\end{tabular}
\end{minipage}
\end{table}

\FloatBarrier

\section{Computational Resources}
\label{app:compute}

The temporal replications and additional objective studies involve
different sets of training runs. The resource estimates below distinguish
measured runtimes from extrapolations to a complete experimental pass.

Training used NVIDIA RTX A5000 GPUs with 24~GB of memory, with four GPUs on one
workstation and, for the block and tuning experiments, a cluster
equipped with the same GPUs. The block
replication trains nine encoders at twenty seeds for the seven
block--source pairs whose training range differs from the mixed-horizon setup, 1{,}260
runs of 100 epochs at batch size 128, one seed per GPU. The median
wall-clock per seed was 0.7 to 1.1 minutes for the six joint-embedding MLP
encoders and under 0.3 minutes for hierarchical contrast, neighbourhood
discrimination and generative reconstruction.
Measured runtimes are available for 960 runs, which took 16.7
GPU-hours. At the same per-encoder means the whole pass is about 20
GPU-hours. The 1{,}040 runs of the context and decomposition methods and their published
objectives, one seed per GPU, took 13.4 GPU-hours. With their evaluation,
tuning, the synthetic study and the held-out families that pass is also
about 20 GPU-hours. On real data seed labels are $\{42, \dots, 61\}$ for
every method, selector and label family, and the synthetic study uses 42 to
46. Encoder and raw rows with the same label share no draw, as defined in
Section~\ref{sec:inference}.

\section{Supporting Research Materials}
\label{app:repro}

The label constructions and fitting dates, temporal splits and
synthetic-data design are documented in this paper. We plan to release
per-seed results, feature representations and trained model parameters.
\Needspace{12\baselineskip}
\section{Data Sources and Access}
\label{app:data}

The interpretation of a transfer score also depends on which assets
enter its window features. The selection rules below specify those
populations and complement the target-universe sensitivity analysis.

\subsection{Asset Selection}
\label{app:data_selection}
The crypto candidate pool contains fifty USDT-quoted Binance spot pairs
with long histories and mean daily quote volume above one million USDT
over 2019--2024. Window features use the twenty pairs with the longest
history, excluding stablecoin pairs. Chronos reads the full pool, which
includes two stablecoin pairs.

The equity panel starts from 225 large S\&P 500 constituents ranked by
early-2025 capitalisation. Among the 191 names with complete histories,
the selection keeps the first fifty after a coverage tie. This produces
an almost alphabetical panel from A to D, rather than a size-ranked
sample. Appendix~\ref{app:universe} evaluates alternative target panels.

The FX universe contains the thirty best-covered of 42 G10 crosses and
emerging-market dollar pairs. Coverage ties also affect the final places.
The additional commodity, REIT and bond families use ten ETFs each.
All universes use later survival or capitalisation information and
therefore inherit retrospective selection effects.

\subsection{Data Access}

Crypto bars come from Binance public data. Equity, FX, ETF and VIX
series come from Yahoo Finance. The Sahm-rule indicator
\texttt{SAHMREALTIME} and recession indicator \texttt{USREC} come
from FRED.
Binance and FRED data are public. Yahoo Finance data are accessed for
personal, non-commercial research and are not redistributed. We plan to
release derived result aggregates, feature matrices, window-date
indices, splits and label arrays under CC BY 4.0, with a datasheet and
protocol cards. Monthly \texttt{SAHMREALTIME} values are assigned from
the fifteenth day after their reference month, following publication.
\texttt{USREC} is dated with hindsight and is not scored.

\end{document}